# Once Correct, Still Wrong: Counterfactual Hallucination in Multilingual Vision-Language Models

**Basel Mousi**♠ **Fahim Dalvi** **Shammur Chowdhury** **Firoj Alam** **Nadir Durrani**♠
Qatar Computing Research Institute, HBKU, Doha, Qatar
{bmousi,faimaduddin,shchowdhury,fialam,ndurrani}@hbku.edu.qa

## Abstract

Vision–language models (VLMs) can achieve high accuracy while still accepting *culturally plausible but visually incorrect* interpretations. Existing hallucination benchmarks rarely test this failure mode, particularly outside Western contexts and English. We introduce **M**$^2$**CQA**, a culturally grounded multimodal benchmark built from images spanning 17 MENA countries, paired with contrastive true and counterfactual statements in English, Arabic, and its dialects. To isolate hallucination beyond raw accuracy, we propose the **CounterFactual Hallucination Rate (CFHR)**, which measures counterfactual acceptance conditioned on correctly answering the true statement. Evaluating state-of-the-art VLMs under multiple prompting strategies, we find that CFHR rises sharply in Arabic, especially in dialects, even when true-statement accuracy remains high. Moreover, reasoning-first prompting consistently increases counterfactual hallucination, while answering before justifying improves robustness. We make the dataset publicly available for the community.[1]

## 1 Introduction

Hallucination in vision-language models (VLMs) is increasingly recognized as a major limitation for deploying multimodal AI systems in real-world environments (Fu et al., 2023; Chen et al., 2025; Li et al., 2025). However, current hallucination benchmarks largely lack **culturally diverse imagery**, and **region-specific visual reasoning**. Models are typically evaluated on datasets dominated by western-centric visual contexts (Romero et al., 2024), limiting our ability to diagnose hallucinations that arise specifically from cultural priors or under-representation. This gap is especially pronounced in the Middle East and North Africa (MENA), a region with rich visual identities, distinctive architectural styles, traditional clothing, regional cuisines, market scenes, and linguistic cues, that are poorly represented in mainstream multimodal corpora (Al-Khalifa et al., 2025). As a result, VLMs frequently misidentify region-specific objects, conflate cultural artifacts across neighboring countries, or default to globally common priors rather than relying on the actual image content.

Beyond the visual domain, hallucination in multilingual and low-resource settings poses an additional challenge. Arabic comprises multiple dialects with substantial lexical and morphological variation, yet dialectal varieties remain severely underrepresented in modern text and multimodal resources (Bouamor et al., 2018; Mousi et al., 2025; Alwajih et al., 2024, 2025). This resource imbalance can cause vision–language models to misinterpret dialectal prompts and hallucinate image-inconsistent answers, particularly when dialect-specific vocabulary is used. However, existing hallucination benchmarks rarely evaluate multimodal behavior across dialects or examine how linguistic uncertainty interacts with visual grounding.

To address these gaps, we introduce **M**$^2$**CQA** (*Multimodal MENA Contrastive Question Answering*), a culturally grounded multimodal benchmark consisting of images from 17 MENA countries paired with contrastive true and counterfactual statements. In the taxonomy of multimodal hallucination evaluation, M$^2$CQA is a *discriminative* benchmark that targets *faithfulness* hallucination. It tests whether models accept image-grounded true statements while rejecting culturally plausible but visually unsupported counterfactuals (Chen et al., 2025). M$^2$CQA systematically evaluates whether VLMs can recognize region-specific visual semantics, rely on the image rather than memorized priors, and remain robust across multiple Arabic dialects of varying resource availability.

At the core of our benchmark is a contrastive true-false design, pairing each image with one grounded true statement and multiple *culturally plausible but visually incorrect counterfactuals*. To study the role of language in hallucination, each question set is constructed in **English, Modern Standard Arabic (MSA)**, and two major Arabic dialects, **Egyptian** and **Levantine**, enabling analysis of how multilinguality and low resourcedness affect a model's ability to distinguish true visual evidence from culturally tempting distractors.

♠ The authors contributed equally.

[1] https://huggingface.co/datasets/QCRI/M2CQA

This contrastive setup allows us to move beyond raw accuracy and introduce a more diagnostic metric of hallucination. Simply answering the true statement ($Q^+$) correctly does not guarantee that the model actually relied on the image; it may have guessed or defaulted to memorized priors. Likewise, accepting a counterfactual statement ($Q^-$) is only meaningful if the model had already demonstrated that it can identify the correct interpretation of the image. We therefore define the **Counter-Factual Hallucination Rate (CFHR)**: the rate at which a model accepts a false but culturally plausible counterfactual after correctly answering its paired true statement. CFHR asks a more revealing question: *once a model appears to understand the image, does it still fall for plausible but incorrect alternatives?* By evaluating counterfactual errors only in cases where $Q^+$ was correct, CFHR isolates failures of visual grounding and cultural reasoning, rather than conflating them with missing background knowledge.

To understand how multimodal systems behave under culturally grounded counterfactual evaluation, we study state-of-the-art vision language models, including **Qwen-VL** (Bai et al., 2023), **Gemma-VL** (Team et al., 2025a), and Arabic focused models such as **Fanar-Oryx** (Team et al., 2026) and **AIN** (Heakl et al., 2025a), and evaluate them under three prompting strategies that vary in structure, justification requirements, and reasoning. Our analysis is guided by five research questions:

**RQ1: Do standard accuracy metrics adequately capture hallucination?**
**Finding:** While $Q^+$ accuracy, $Q^-$ accuracy, and F1 summarize performance, they often mask cases where models answer $Q^+$ correctly yet still accept culturally plausible counterfactuals. CFHR reveals these conditional grounding failures that are systematically obscured by aggregate metrics.

**RQ2: How does hallucination vary across languages and dialects?**
**Finding:** CFHR increases substantially from English to MSA and rises further in dialectal Arabic, even when $Q^+$ accuracy remains high. This suggests that linguistic uncertainty amplifies susceptibility to culturally tempting distractors.

**RQ3: How do model families differ in robustness to hallucination?**
**Finding:** Qwen3-VL exhibits consistently lower CFHR across languages than Gemma-3-VL. Arabic-focused models exhibit distinct behavioral patterns. Fanar shows lower CFHR alongside reduced $Q^+$ accuracy, indicating more frequent rejection of both true and counterfactual statements, whereas AIN maintains high $Q^+$ accuracy but exhibits very high CFHR, reflecting a tendency to accept culturally plausible counterfactuals even when the true statement is correctly identified.

**RQ4: How does prompting strategy affect hallucination?**
**Finding:** Requiring evidence *after* committing to an answer (*Answer then Reason*) generally reduces CFHR, while eliciting reasoning *before* the final decision (*Reasoning First*) often increases CFHR, especially in Arabic and its dialects.

**RQ5: Does model scaling reduce hallucination?**
**Finding:** Scaling generally lowers CFHR, particularly in Arabic and dialectal settings, but the gains are model-family dependent. Qwen3-VL shows stronger and more consistent reductions with scale, whereas Gemma-3-VL improvements are smaller and tend to saturate at larger sizes.

We make **three major contributions**:

- We introduce **M²CQA**, a culturally grounded multimodal benchmark built from images spanning 17 MENA countries, paired with contrastive true statements and culturally plausible counterfactuals across English, MSA, and Arabic dialects.
- We propose the **CounterFactual Hallucination Rate (CFHR)**, a conditional metric that isolates counterfactual hallucination by measuring false acceptance after a model correctly identifies the image-grounded true statement.
- We provide a systematic evaluation of state-of-the-art VLMs, including **Qwen3-VL**, **Gemma-3-VL**, and Arabic-focused models (**Fanar**, **AIN**), across prompting strategies and model scales, revealing substantial differences in cultural robustness and a strong sensitivity to reasoning order.

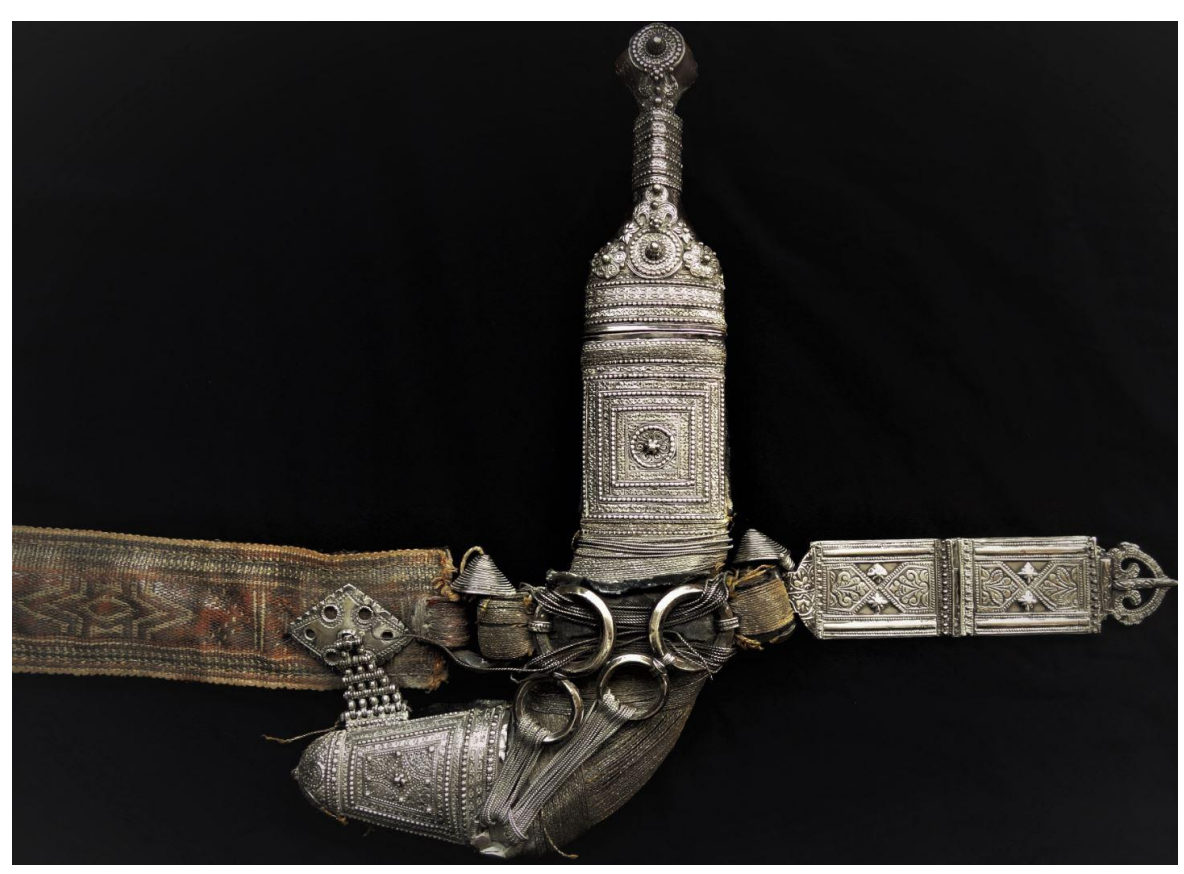

Figure 1: Sample image from the M$^2$CQA dataset.
**Q**$^+$ The weapon in the image is a **traditional khanjar**.
**Q**$^-$ ... a **ceremonial sword** used in traditional dances.
**Q**$^-$ ... a **hunting knife** designed for outdoor survival.

While our experiments focus on the MENA region, the proposed CFHR metric and counterfactual construction framework are region-agnostic and can be readily applied to other cultural settings.

## 2 M$^2$CQA Curation

We construct **M$^2$CQA**, a culturally grounded multimodal benchmark sourced from an in-house data collection. The dataset covers a broad range of culturally distinctive visual themes, including architecture, markets, traditional clothing, public squares, historical sites, and region-specific artifacts, providing a rich basis for evaluating visual understanding across culturally diverse settings.

**Image Collection and Curation:** Images were collected using the Google Custom Search (Image) API with geo-targeted, country-specific queries derived from a taxonomy of visual categories, following the structured retrieval framework of EverydayMMQA (Alam et al., 2025). The taxonomy defines 12 top-level categories to ensure broad coverage of culturally salient visual patterns in MENA (Middle East and North Africa), including built environments, food, clothing, religion, daily activities, transport, and natural scenes. Queries were manually refined to ensure geographic relevance (e.g., demonyms and local terminology). We restricted retrieval to Creative Commons licensed images (`cc_publicdomain`, `cc_attribute`, `cc_sharealike`) and enabled SafeSearch. For each query, we collected top-ranked results with caps to maintain balanced coverage across countries and categories. Images were filtered through manual screening and automatic deduplication, including exact URL matching and near-duplicate removal following prior dataset curation practice (Alam et al., 2024). The final dataThis obs set spans 17 countries; see Appendix A.1 for taxonomy and dataset details.

**MCQ-Based Annotation and Filtering:** For each image, we first construct a multiple-choice question (MCQ) with one correct option and several plausible distractors. These MCQs are generated using GPT-4.1 and serve only as an intermediate annotation scaffold to enumerate culturally plausible interpretations of the image. Before constructing true/false statements, we apply a visual-grounding filter to reduce the influence of language priors. Concretely, we evaluate each candidate MCQ in an image-blind (text-only) setting using GPT-4.1 and Gemini-2.5-Flash, and discard QAs that both models answer correctly without access to the image, as they are more likely answerable from language cues alone rather than visual evidence.

**Counterfactual Statement Curation:** From each retained MCQ, we derive a contrastive true/-false set consisting of one true statement (**Q**$^+$) corresponding to the correct MCQ option and two false but culturally plausible counterfactual statements (**Q**$^-$) derived from the incorrect MCQ options. This conversion is performed using GPT-4.1, which rewrites each MCQ option into a declarative statement while preserving its semantic content. By construction, the resulting counterfactuals are minimally contrastive with the true statement, differing along fine-grained cultural or visual dimensions. In practice, these variations follow structured semantic patterns, including landmark or national misattribution, culturally similar event confusions, cuisine or attire substitutions, and visually similar object or scene-type shifts within the same broader category. These substitutions typically preserve the overall scene context while altering a fine-grained cultural, semantic, or identity attribute, while remaining visually unsupported by the image, as illustrated in Figure 1. The final benchmark evaluates each statement independently as a binary true/false judgment. Please see Appendix A.2 for curation prompts and Appendix A.8 for additional samples.

**Machine Translation:** MCQs are translated into Modern Standard Arabic (MSA) using the in-house **Fanar Shaheen** machine translation system (Team et al., 2026) and into dialectal Arabic using GPT-

| Statement | Need-Image | | Q-Correctness | |
|---|---|---|---|---|
| | AC1 | Raw | AC1 | Raw |
| $Q^+$ | 0.996 | 0.996 | 0.752 | 0.784 |
| $Q_1^-$ | 0.997 | 0.997 | 0.775 | 0.803 |
| $Q_2^-$ | 0.996 | 0.996 | 0.695 | 0.738 |
| **Avg.** | 0.996 | 0.996 | 0.741 | 0.775 |

Table 1: Inter-annotator agreement for *Need-Image* and *Q-Correctness*.

4.1, then converted into contrastive true/false statements following the same procedure as in English to preserve semantic equivalence across languages (see Appendix A.4 for MT evaluation).

**Human Verification and Quality Control:** We assess annotation reliability through a human verification study on a randomly sampled subset comprising 15% of the dataset. The subset is stratified to ensure proportional coverage across countries and taxonomic categories. Annotators are shown an image and three associated question–answer (Q/A) pairs, and assign two independent labels: *(i) Need-Image*, indicating whether answering the statement requires access to the image (*Yes*, *No*, *Unsure*), and *(ii) Q-Correctness*, indicating whether the statement is visually supported by the image (*Correct*, *Incorrect*, *Unsure*). Answers lacking clear visual support, relying on external knowledge, or irrelevant to the image are marked *Incorrect*, while *Unsure* is reserved for genuine visual ambiguity.

Each item is annotated by two annotators, with a third annotator resolving disagreements. We report both Gwets AC1 (Gwet, 2008) and raw agreement. Table 1 summarizes the results. Agreement is near-perfect for *Need-Image* (AC1 $\approx$ 0.996), indicating consistent identification of visually grounded questions. For *Q-Correctness*, agreement is substantial overall (AC1 $\approx$ 0.741; raw $\approx$ 0.775), reflecting the finer-grained judgments required for contrastive evaluation. Bootstrap 95% confidence intervals (item-level resampling) are strictly above zero across all conditions. Qualitative inspection indicates that disagreements primarily arise from visually subtle counterfactual distinctions or fine-grained cultural ambiguity, rather than systematic annotation errors. Instances of image insufficiency are rare, consistent with the high agreement for *Need-Image*. Full annotation guidelines and interface details are provided in Appendix A.3.

**$M^2$CQA Dataset:** The final $M^2$CQA dataset consists of 9,990 visually grounded samples, each comprising one image paired with one true statement and two culturally plausible counterfactuals, instantiated across multiple language varieties. See Appendix A.1 for country-wise distribution and taxonomy details. This dataset enables systematic evaluation of multimodal reasoning across languages and culturally grounded contexts.

## 3 CounterFactual Hallucination Rate

Accuracy on the true statement ($Q^+$) does not reveal whether a model is genuinely grounded in the image or merely relying on coarse cultural priors. A model may answer $Q^+$ correctly by associating lanterns with generic festive contexts, without attending to the specific visual evidence that distinguishes different celebrations. As shown in Figure 2, one image places a decorative lantern within a street scene characterized by architectural structure and ambient lighting patterns commonly associated with Ramadan settings, whereas the other depicts a lantern embedded in a winter scene with visible snow and seasonal decorations such as baubles and evergreen branches, consistent with a Christmas context. A model that is not visually grounded may correctly identify one image as representing Ramadan while still accepting counterfactual interpretations that attribute the same scene to Christmas or Diwali, or may fail to distinguish it from the visually similar Christmas lantern image. By contrast, a visually grounded model would rely on these contextual cues to correctly identify the first image as representing Ramadan and the second as representing Christmas, while rejecting culturally plausible but visually unsupported alternatives. This joint success on $Q^+$ and $Q^-$ provides a stronger diagnostic of visual grounding than $Q^+$ accuracy alone.

To better capture this behavior, we condition hallucination on the models ability to answer $Q^+$ correctly. Errors on counterfactual statements ($Q^-$) are only meaningful if the model first succeeds on $Q^+$. If the model fails $Q^+$ due to missing knowledge or misunderstanding, its behavior on $Q^-$ does not constitute hallucination. The relevant question is therefore: *once a model appears to understand the image, does it still fall for plausible but incorrect alternatives?* Conditioning hallucination on $Q^+$ allows us to isolate failures of visual grounding and cultural reasoning from errors that stem merely

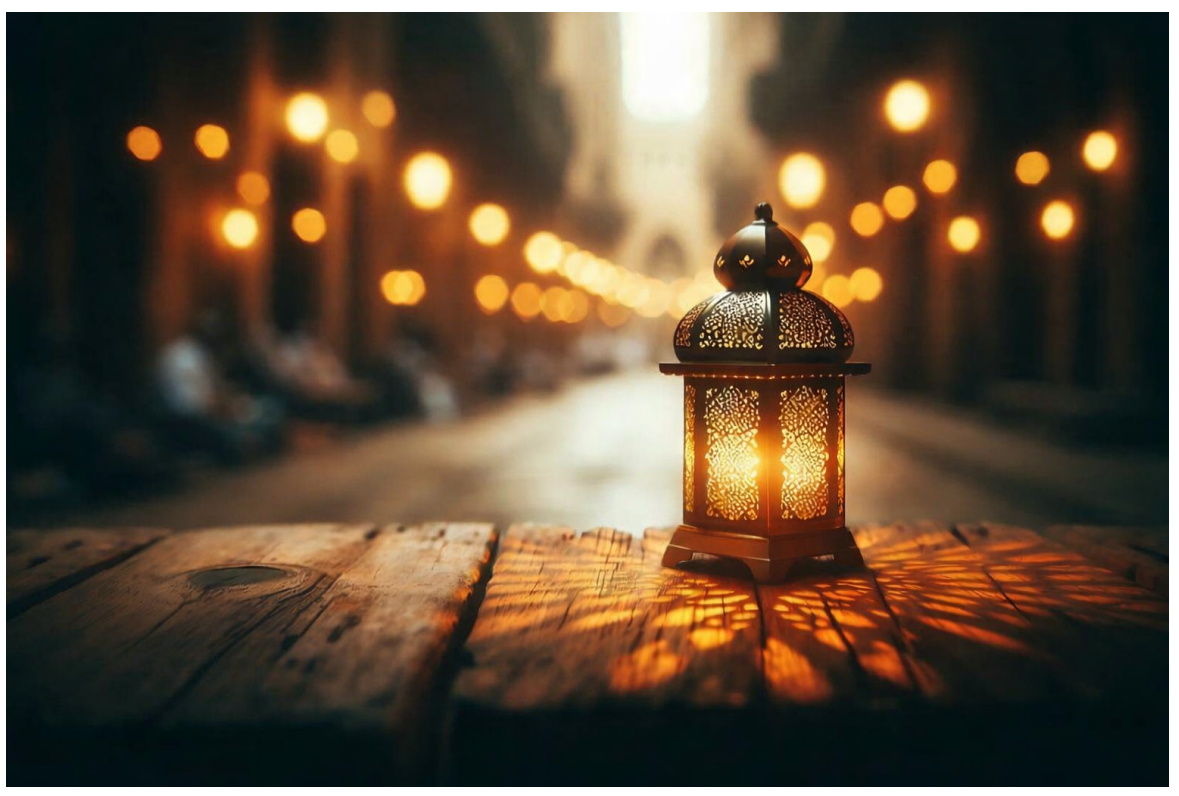

**Q**$^+$ The decorative lantern and warm lights are most likely representing **Ramadan celebrations**.
**Q**$^-$ ... representing **Christmas Eve**.
**Q**$^-$ ... representing the **Diwali festival**.

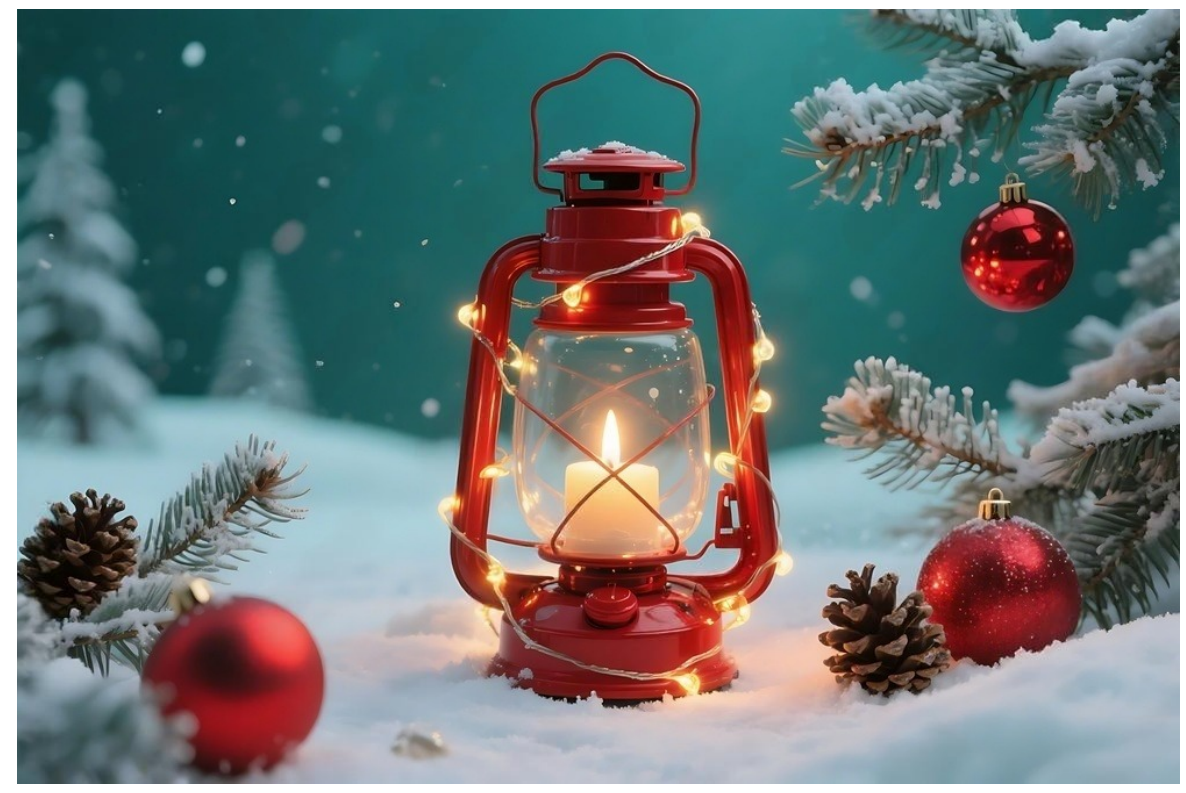

**Q**$^+$ The decorative lantern and warm lights are most likely representing **Christmas Eve**.
**Q**$^-$ ... representing **Ramadan celebrations**.
**Q**$^-$ ... representing the **Diwali festival**.

Figure 2: Visually similar lantern scenes with distinct cultural grounding. A grounded model should answer Q$^+$ correctly and reject culturally plausible but visually unsupported alternatives (Q$^-$).

from incomplete knowledge.

Each item in M$^2$CQA consists of one grounded true statement (Q$^+$) and two culturally plausible false statements (Q$^-$). A model demonstrates correct contrastive grounding only when it answers Q$^+$ correctly and rejects both counterfactuals. Let $\text{Acc}(Q^+)$ denote accuracy on Q$^+$ alone, and let $\text{Acc}(\text{combined})$ denote accuracy on answering both Q$^+$ and all Q$^-$ statements correctly. The difference between these quantities captures the proportion of cases in which a model recognizes the true statement but still accepts at least one counterfactual alternative. We define the **CounterFactual Hallucination Rate (CFHR)** as:

$$
\begin{aligned}
\text{CFHR} &= \frac{\text{Acc}(Q^+) - \text{Acc}(\text{combined})}{\text{Acc}(Q^+)} \\
&= P(\text{fails on any Q}^- \mid \text{succeeded on Q}^+)
\end{aligned}
$$

This conditional formulation isolates hallucination that occurs despite the model identifying the correct visual grounding. High CFHR values indicate that a model defaults to memorized regional stereotypes or culturally familiar associations even when it has already demonstrated recognition of the true interpretation. CFHR therefore provides a more diagnostic measure of hallucination in culturally rich, visually grounded settings than raw accuracy alone.

Contemporary metrics such as POPE (Li et al., 2023), CHAIR (Rohrbach et al., 2018), and object hallucination rate (Liu et al., 2024) measure hallucination by treating each sample independently, overlooking overall model weaknesses. For instance, POPE uses binary QA (Yes/No) to probe object presence and relies on task accuracy as a hallucination measure, while CHAIR evaluates object hallucination in image captioning via passive observation. In contrast, CFHR introduces a conditional dependency on success on both Q$^+$ and Q$^-$, forcing models to adjudicate between visually grounded truth and cultural plausibility. This provides a more rigorous diagnostic for models deployed in culturally rich and visually complex environments.

As a conditional metric, CFHR is not intended to be interpreted in isolation. In particular, low CFHR does not imply strong visual grounding, as models may exhibit low CFHR by rarely satisfying the conditioning event (i.e., correctly answering Q$^+$). CFHR should therefore be interpreted with respect to its conditioning variable, distinguishing counterfactual hallucination from reduced coverage or conservative rejection behavior. We revisit this interaction in the results.

## 4 Results

**Setup:** We evaluate a range of vision language models spanning different families and scales: Qwen3-VL (2B, 4B, 8B, 32B) (Bai et al., 2025), Gemma-3 (4B, 12B, 27B) (Team et al., 2025b), Fanar-Oryx (Fanar-Team et al., 2025), and AIN (Heakl et al., 2025b), using the `vLLM` inference li-brary.[2] Models are evaluated under three prompting settings: *True or False*, *Answer then Reason*, and *Reasoning First*, and across four language variants: English, MSA, Levantine, and Egyptian. Full prompt templates and output formats are provided in Appendix A.5. Table 2 summarizes the performance of multiple vision-language models across languages and dialects under three prompting strategies. Based on these results, we address the following research questions.

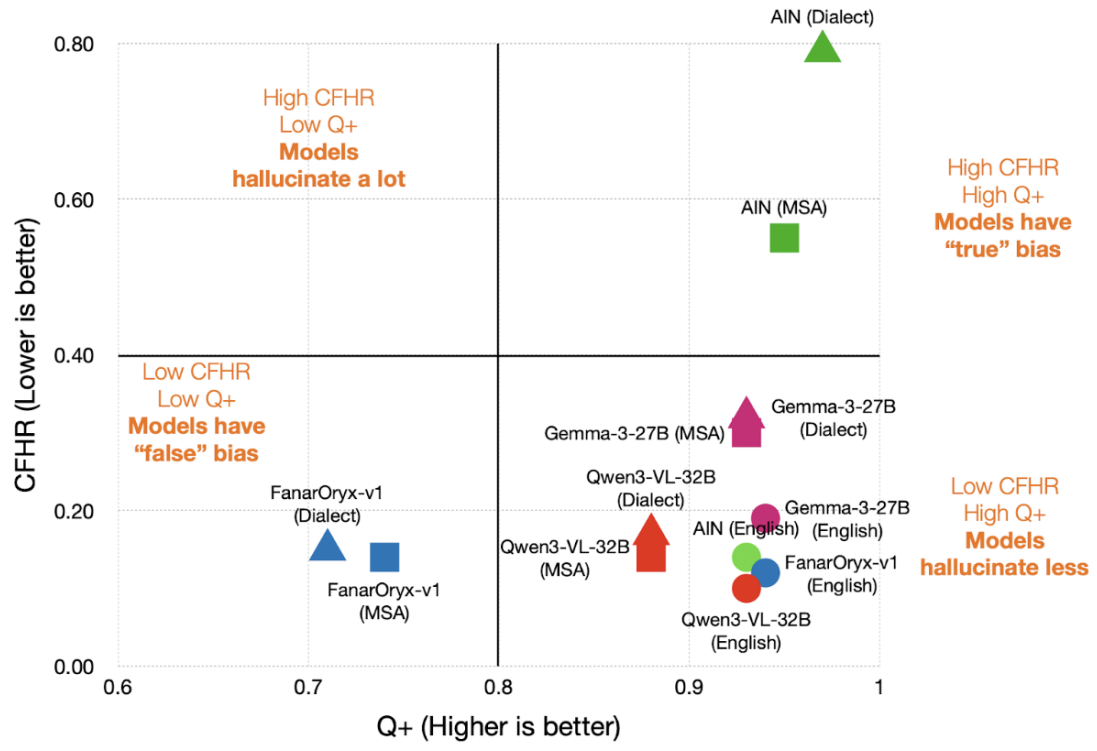

Figure 3: Joint visualization of $Q^+$ accuracy and CFHR. The two-dimensional view reveals distinct behavioral regimes: models with high $Q^+$ and low CFHR exhibit strong grounding and counterfactual robustness, while other regions reflect conservative rejection (low $Q^+$, low CFHR), over-acceptance (high $Q^+$, high CFHR), or frequent hallucination (low $Q^+$, high CFHR).

**RQ1: Do existing metrics adequately capture hallucination in multimodal reasoning?** While $Q^+$, $Q^-$, and F1 summarize overall accuracy, they do not reveal whether a model can reject plausible but incorrect counterfactual interpretations. As shown in Table 2, models often achieve high accuracy on true statements ($Q^+$) and competitive F1 scores, yet still accept culturally plausible but visually incorrect counterfactuals after identifying the correct interpretation. This behavior arises naturally in our true counterfactual T/F setup, where false statements are intentionally constructed to align with regional priors rather than random negatives. For instance, while AIN on dialectal Arabic attains a near-ceiling $Q^+$ of 0.97 yet exhibits a CFHR of 0.79, indicating frequent acceptance of culturally plausible counterfactuals, this pattern is not limited to degenerate behaviors, as Gemma-3-27B in MSA achieves a strong F1 of 0.86 ($Q^+$ = 0.93) while still incurring a CFHR of 0.30. The proposed CFHR isolates this conditional failure by measuring hallucination only in cases where the model has already succeeded on $Q^+$. CFHR therefore provides a complementary diagnostic perspective that is particularly well suited to culturally grounded multimodal evaluation, revealing grounding failures that are obscured by standard accuracy metrics. As a conditional metric, CFHR should be interpreted with respect to $Q^+$ accuracy, as low CFHR may also arise from conservative rejection behavior rather than strong visual grounding.

To aid interpretation, we visualize model behavior jointly in terms of $Q^+$ accuracy and CFHR (Figure 3), which reveals distinct behavioral regimes. Models with high $Q^+$ and low CFHR exhibit both strong visual grounding and counterfactual robustness, while low CFHR paired with low $Q^+$ often reflects conservative rejection behavior rather than genuine grounding. Conversely, high CFHR despite strong $Q^+$ indicates over-acceptance of plausible but unsupported alternatives. This joint view highlights that CFHR should not be interpreted in isolation, but in conjunction with $Q^+$ accuracy.

**RQ2: How does counterfactual hallucination vary across languages and dialects?** Table 2 shows a consistent language-dependent trend across all models. While performance in English exhibits low counterfactual hallucination rates, CFHR increases substantially in MSA and rises further in dialectal Arabic. Notably, this increase occurs even when $Q^+$ accuracy remains relatively high, indicating that models often identify the correct interpretation but fail to reliably reject counterfactual alternatives under increased linguistic uncertainty. In contrast, F1 scores degrade only moderately across languages, suggesting that aggregate accuracy metrics substantially underestimate the severity of hallucination in lower-resource and dialectal settings. We note that this relationship should be interpreted as correlational rather than causal. The observed pattern of increasing CFHR alongside decreasing Q+ accuracy suggests that reduced grounding reliability may contribute to higher susceptibility to culturally plausible distractors, though alternative factors such as dialectal data sparsity or distributional mismatch may also play a role.

**RQ3: How do different VLMs differ in their robustness to counterfactual hallucination?** Table 2 reveals substantial variation in counterfactual hallucination behavior across model families. Qwen3-VL consistently exhibits the strongest ro-bustness, maintaining low CFHR across languages and showing modest improvements under evidence-based prompting. In contrast, Gemma-3 displays higher CFHR in MSA and dialectal Arabic, with increased model scale offering limited mitigation.

[2] https://docs.vllm.ai/en/latest/

| Lang | Model | A. True or False | | | | B. Answer then Reason | | | | C. Reasoning First | | | |
|---|---|---|---|---|---|---|---|---|---|---|---|---|---|
| | | $Q^+$ | $Q^-$ | F1 | **CFHR** $\downarrow$ | $Q^+$ | $Q^-$ | F1 | **CFHR** $\downarrow$ | $Q^+$ | $Q^-$ | F1 | **CFHR** $\downarrow$ |
| EN | Qwen3-VL-32B | 0.93 | 0.94 | 0.93 | **0.10** | 0.91 | 0.95 | 0.93 | **0.08** | 0.93 | 0.93 | 0.93 | **0.12** |
| | Gemma-3-27B | 0.94 | 0.88 | 0.91 | **0.19** | 0.92 | 0.89 | 0.90 | **0.18** | 0.92 | 0.90 | 0.91 | **0.16** |
| | FanarOryx-v1 | 0.94 | 0.93 | 0.93 | **0.12** | 0.94 | 0.93 | 0.93 | **0.12** | 0.91 | 0.92 | 0.92 | **0.13** |
| | AIN | 0.93 | 0.91 | 0.92 | **0.14** | 0.88 | 0.93 | 0.90 | **0.11** | 0.85 | 0.84 | 0.85 | **0.25** |
| MSA | Qwen3-VL-32B | 0.88 | 0.91 | 0.90 | **0.14** | 0.85 | 0.93 | 0.89 | **0.11** | 0.86 | 0.91 | 0.89 | **0.15** |
| | Gemma-3-27B | 0.93 | 0.81 | 0.86 | **0.30** | 0.92 | 0.83 | 0.87 | **0.27** | 0.90 | 0.86 | 0.88 | **0.23** |
| | FanarOryx-v1 | 0.74 | 0.92 | 0.82 | **0.14** | 0.64 | 0.95 | 0.76 | **0.11** | 0.77 | 0.78 | 0.78 | **0.34** |
| | AIN | 0.95 | 0.60 | 0.74 | **0.55** | 0.92 | 0.77 | 0.84 | **0.35** | 0.71 | 0.54 | 0.61 | **0.66** |
| Dialect | Qwen3-VL-32B | 0.88 | 0.90 | 0.88 | **0.17** | 0.85 | 0.92 | 0.88 | **0.14** | 0.87 | 0.90 | 0.88 | **0.17** |
| | Gemma-3-27B | 0.93 | 0.79 | 0.86 | **0.32** | 0.93 | 0.81 | 0.86 | **0.31** | 0.91 | 0.85 | 0.88 | **0.24** |
| | FanarOryx-v1 | 0.71 | 0.92 | 0.80 | **0.15** | 0.59 | 0.95 | 0.73 | **0.11** | 0.76 | 0.74 | 0.75 | **0.39** |
| | AIN | 0.97 | 0.38 | 0.56 | **0.79** | 0.91 | 0.72 | 0.80 | **0.43** | 0.65 | 0.58 | 0.61 | **0.63** |

Table 2: Comparison of prompting strategies across languages. *Answer only* uses direct answer prediction, *Answer then Reason* requires a final answer followed by justification, and *Reasoning First* elicits reasoning prior to committing to an answer. CFHR denotes the Counterfactual Hallucination Rate, measuring hallucination conditional on correctly answering $Q^+$. Dialectal results are averaged over Egyptian and Levantine Arabic.

Fanar and AIN, both Qwen-based models further trained with a focus on Arabic language and culture, exhibit distinct but informative failure modes. Fanar adopts a conservative strategy, achieving relatively low CFHR across Arabic settings at the cost of reduced $Q^+$ accuracy, particularly in lower-resource languages. AIN, by contrast, attains very high $Q^+$ accuracy in Arabic and dialectal settings while exhibiting extremely high CFHR, indicating a tendency to accept culturally plausible counterfactual statements even when the true statement is correctly identified. This pattern suggests that stronger cultural alignment and language specialization do not necessarily translate into improved robustness against counterfactual hallucination.

**RQ4: How does evidence elicitation affect counterfactual hallucination?** Prior work suggests that model behavior can vary substantially depending on when reasoning or evidence is elicited (Wei et al., 2022; Kojima et al., 2022; Lampinen et al., 2022). In Table 2, we compare three evaluation settings: direct true/false prediction (*True or False*), answering followed by evidence (*Answer then Reason*), and reasoning before committing to an answer (*Reasoning First*).

First, comparing *True or False* (A) with *Answer then Reason* (B), we observe a consistent reduction in CFHR across languages and models, indicating that requiring evidence *after* committing to an answer improves robustness to counterfactuals. This effect is most visible in MSA and dialectal Arabic (e.g., AIN: 0.55→0.35 in MSA; 0.79→0.43 in dialects), suggesting that post-hoc justification encourages models to ground their decisions explicitly in the image rather than relying on culturally plausible but visually unsupported associations.

Second, comparing *Answer then Reason* (B) with *Reasoning First* (C), we observe the opposite pattern. CFHR frequently increases when models are asked to produce reasoning *before* committing to a final decision. This effect is particularly pronounced in Arabic varieties, where several models exhibit substantially higher CFHR under *Reasoning First* than under *Answer then Reason*. For example, models such as AIN in MSA and Fanar in dialectal Arabic show marked increases in counterfactual hallucination, indicating that eliciting reasoning prior to answering can significantly exacerbate hallucinations relative to (B).

A plausible explanation is that reasoning before commitment encourages models to explore alternative interpretations and, in doing so, to introduce additional semantic or cultural constraints that effectively reshape the decision criterion of the task. These constraints are often not required by the benchmark and are not supported by the visual evidence. In contrast, committing to an answer first may "anchor" the model to an image-grounded interpretation of the question, making subsequent evidence generation more likely to rationalize a perceptual judgment rather than drift toward culturally,

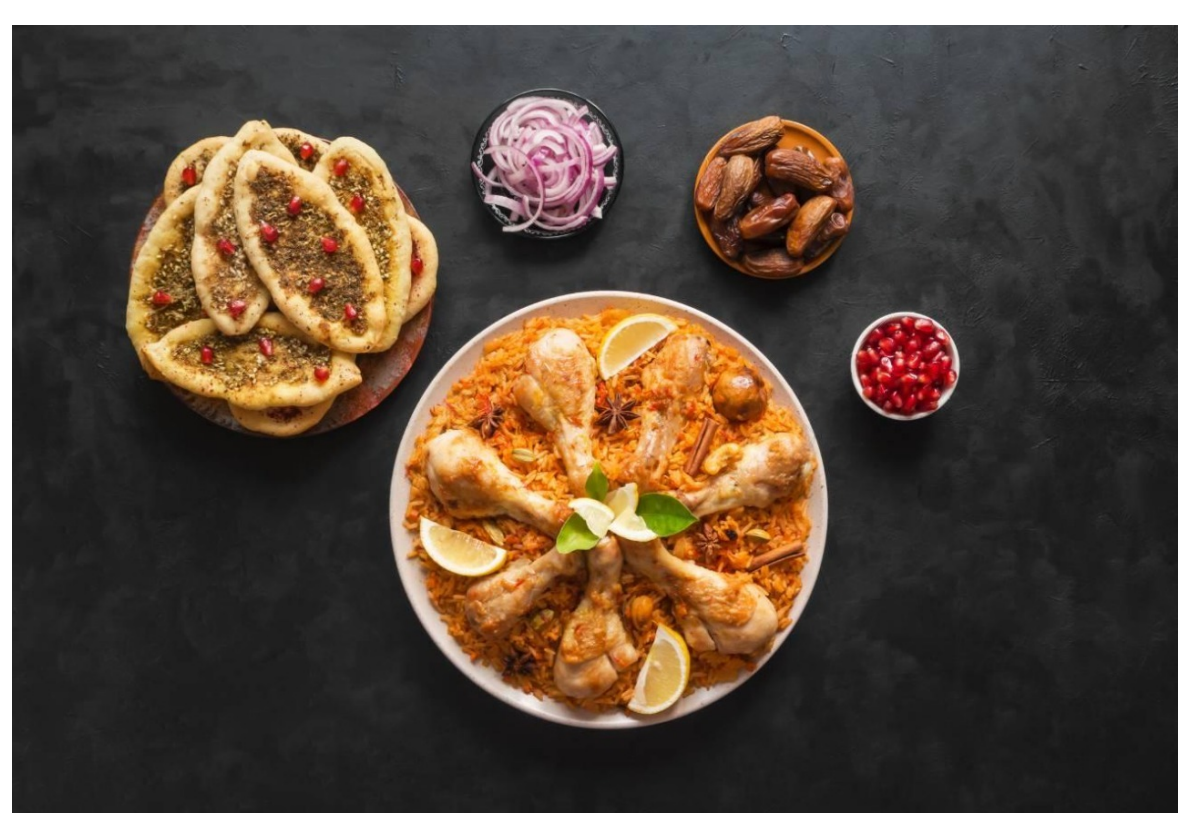

Figure 4: **Reasoning-first induces counterfactual rejection.** Example illustrating the effect of prompting strategy on model behavior. **Question:** Star anise is commonly used in Middle Eastern cuisine and is visible in the chicken biryani shown in the image. True or False? **Ground Truth:** True (star anise is visible). **Prompt B (Answer then Reason):** *Answer: True.* The model correctly judges that star anise is present in the chicken biryani, grounding its decision in visible evidence. **Prompt C (Reasoning First):** *Answer: False.* Although star anise is visible, the model reasons that it is not *commonly* used in Middle Eastern cuisine and is more prevalent in South and East Asian cooking, shifting the decision criterion from visual presence to culinary typicality and overriding an image-grounded judgment. See Table 9 in Appendix A.6 for details.

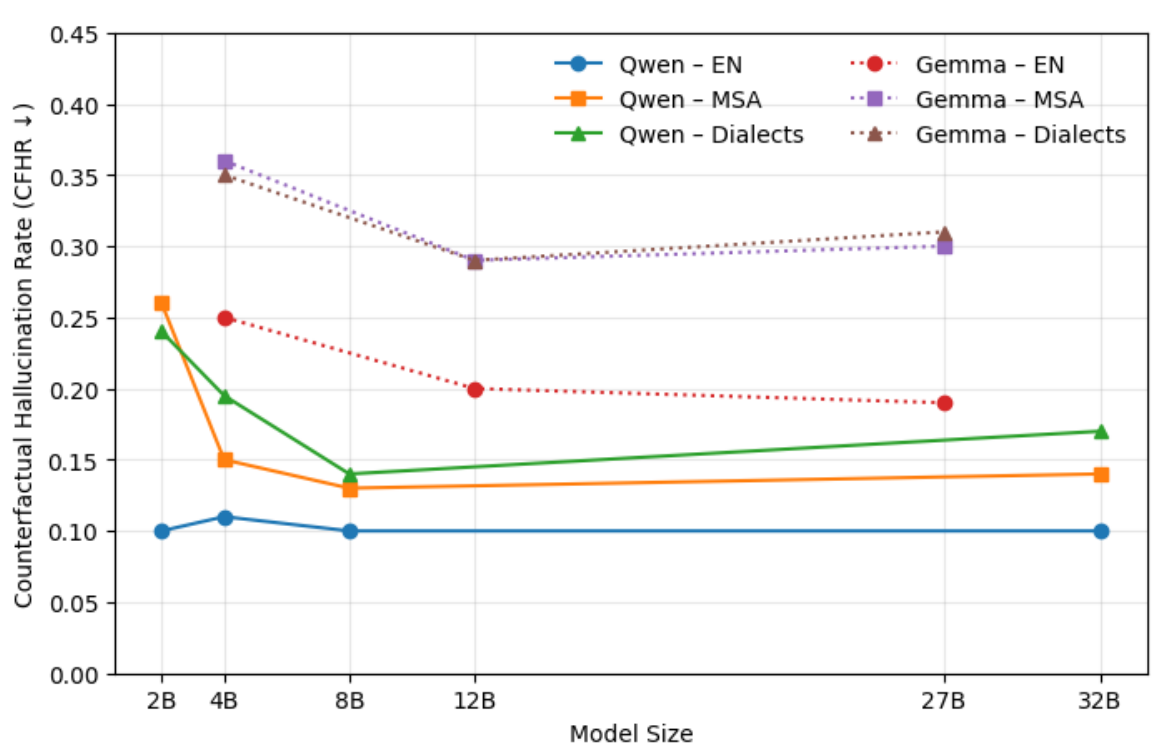

Figure 5: **Scaling and counterfactual hallucination under Prompt A** CFHR (lower is better) as a function of model size for Qwen3-VL (solid) and Gemma-3-VL (dotted), evaluated in English (EN), Modern Standard Arabic (MSA), and dialectal Arabic (Dialects; average of Egyptian and Levantine). Scaling generally reduces CFHR, with stronger and more consistent improvements for Qwen3-VL, particularly in Arabic varieties.

logically stricter, or over-interpreted alternatives.

This behavior is illustrated in Figure 4 using outputs from the Qwen3-VL-32B model. In this example, the statement reads: *"Star anise is commonly used in Middle Eastern cuisine and is visible in the chicken biryani shown in the image."* Under *Answer then Reason* (Prompt B), the model correctly predicts `True`, grounding its decision in the clearly visible star anise pods present in the dish. Under *Reasoning First* (Prompt C), however, the same model predicts `False` after acknowledging the visual presence of star anise but introducing background knowledge about regional culinary practices, arguing that star anise is not *commonly* used in Middle Eastern cuisine. While this reasoning is culturally informed, it shifts the decision criterion from visual presence to the typicality of ingredient usage across cuisines, introducing an extraneous constraint that is not supported by the image. This example illustrates how pre-answer reasoning can override image-grounded evidence by reframing the task itself, leading to counterfactual errors even when the relevant visual content is correctly perceived. Please see Table 9 in Appendix A.6 for full model reasoning traces and additional qualitative examples illustrating this pattern.

**RQ5: Does model scaling reduce counterfactual hallucination?** Figure 5 shows the effect of model scaling on CFHR under the direct true/false setting (Prompt A). Across both model families, larger models generally exhibit lower counterfactual hallucination, with the effect being most pronounced in Arabic varieties. Qwen3-VL shows consistent reductions in CFHR with scale in MSA and dialectal Arabic, while Gemma-3-VL benefits primarily from scaling up to mid-sized models, after which improvements saturate. Overall, these results confirm the intuitive trend that increased model capacity improves robustness to counterfactual hallucination, while also highlighting substantial differences between model families. The pattern holds for Prompts B and C (Appendix A.7).

## 5 Related Work

Hallucination in vision–language models (VLMs) is categorized into *faithfulness hallucination*, where outputs are inconsistent with the visual input, and *factuality hallucination*, where outputs contradict world knowledge (Chen et al., 2025). Our work focuses on faithfulness hallucination and is most closely related to prior efforts on hallucination benchmarks and evaluation metrics.

### 5.1 Multimodal Hallucination Benchmarks

A large body of work evaluates hallucination through dedicated multimodal benchmarks, most commonly using discriminative task formats such as yes/no questions (YNQs) or MCQs to assess visual grounding. Early benchmarks such as POPE (Li et al., 2023) measure object hallucination by querying object presence in MSCOCO images (Lin et al., 2014), while ROPE (Chen et al., 2024b) and FGHE (Wang et al., 2023) extend this paradigm to multi-object reasoning, attributes, relations, and object behaviors. Other benchmarks explicitly target relational hallucination, including RAH-Bench (Chen et al., 2023) and R-Bench (Wu et al., 2024a), while AutoHallusion (Wu et al., 2024b) automates the construction of object and spatial reasoning probes. More recent datasets such as PhD (Liu et al., 2025) and LongHalQA (Qiu et al., 2024) explore hallucination across diverse visual tasks and long-context settings.

Beyond discriminative benchmarks, generative evaluations assess hallucination in open-ended image-to-text generation. CHAIR (Rohrbach et al., 2018) and its extensions such as OpenCHAIR (Ben-Kish et al., 2024) quantify object hallucination in image captioning by comparing generated captions against ground-truth object annotations, while approaches such as CC-EVAL (Zhai et al., 2024) and NOPE (Lovenia et al., 2024) rely on controlled negative or contrastive generation.

Several comprehensive benchmarks combine discriminative and generative evaluation within a single framework. MME (Fu et al., 2023), AMBER (Wang et al., 2024), and MERLIM (Villa et al., 2025) evaluate multiple hallucination types including object existence, relations, and counting, often by comparing model behavior across original and edited images. Additional benchmarks such as VHtest (Huang et al., 2024), VisDiaHalBench (Cao et al., 2024), Hal-Eval (Jiang et al., 2024), Med-HallMark (Chen et al., 2024a), and ODE (Tu et al., 2025) further expand coverage across domains, task formats, and evaluation protocols.

Despite their breadth, the majority of existing benchmarks are constructed from Western-centric datasets such as MSCOCO and ADE20K and primarily focus on generic object and relation semantics. As a result, they rarely evaluate hallucination in culturally grounded settings or test whether models can distinguish visually supported interpretations from culturally plausible but incorrect alternatives. While CVQA (Romero et al., 2024) introduces culturally diverse images and multilingual question answering, hallucination-specific evaluation in regionally grounded contexts, particularly in the MENA, remains largely unexplored.

### 5.2 Metrics for Multimodal Hallucination

Existing hallucination benchmarks rely primarily on accuracy, F1 score, or unconditional error rates aggregated over all instances. In discriminative settings, hallucination is measured as incorrect responses to negative questions, while generative benchmarks quantify hallucination through object-level precision and recall metrics, as in CHAIR and its variants. Comprehensive benchmarks similarly aggregate performance across heterogeneous tasks using unconditional scores.

A key limitation of these metrics is that they do not condition hallucination on successful visual understanding. Errors due to incomplete knowledge or failure to recognize the image are treated equivalently to cases where a model correctly identifies the true interpretation but still accepts plausible yet unsupported alternatives. Consequently, existing metrics do not explicitly capture whether hallucination persists *after* a model has demonstrated correct visual grounding. This gap motivates conditional evaluation protocols that isolate counterfactual hallucination given successful recognition of image-grounded evidence, which are largely absent from prior multimodal hallucination benchmarks. In contrast, we explicitly condition hallucination on correct identification of the true visual interpretation, enabling the measurement of counterfactual errors that persist even after a model appears to understand the image.

## 6 Conclusion

We introduced **M$^2$CQA** and the **CounterFactual Hallucination Rate (CFHR)** to diagnose counterfactual hallucination in culturally grounded multimodal settings. Across models, hallucination is substantially underestimated by accuracy alone, with CFHR increasing sharply in Arabic and dialectal prompts despite strong performance on true statements. Prompting strategy has a significant effect: committing to an answer before justification reduces hallucination, whereas reasoning-first prompting often amplifies reliance on cultural priors. Our findings highlight the need for culturally grounded evaluation and conditional metrics like CFHR to accurately assess multimodal reasoning under linguistic and cultural uncertainty. We make the dataset publicly available for the community.[3]

[3] https://huggingface.co/datasets/QCRI/M2CQA

## Limitations and Potential Risks

While M$^2$CQA and the CFHR provide a new lens for evaluating counterfactual hallucination in culturally grounded multimodal settings, several limitations remain.

**Geographic and cultural scope:** Although M$^2$CQA spans images from 17 countries across the MENA region, it does not fully capture the cultural, socioeconomic, and visual diversity within each country. The dataset is constructed using a taxonomy-driven, geo-targeted retrieval pipeline designed to provide broad coverage across visual-cultural domains rather than an exhaustive representation of all culturally meaningful concepts. As a result, it may reflect biases inherent to web-based image retrieval, including over-representation of visually salient landmarks, commonly photographed artifacts, or dominant cultural narratives. Intra-country variation (e.g., rural vs. urban settings, regional architectural styles, or minority cultural practices) is not explicitly controlled for, and results are reported in aggregate rather than at the country level. Consequently, observed trends should be interpreted as region-level patterns rather than fine-grained national or subcultural analyses.

**Language and dialect coverage:** Our multilingual evaluation focuses on English, MSA, and two widely spoken Arabic dialects (Egyptian and Levantine). While these dialects cover a large portion of Arabic speakers, other major varieties (e.g., Gulf, Maghrebi, or Iraqi Arabic) are not included. In addition, dialectal statements are generated via machine translation rather than original human authoring, which may introduce translation artifacts despite quality checks. These factors may limit the generalizability of dialect-specific findings.

**Scale of human verification:** Human verification was conducted on a limited subset of the dataset due to budget constraints. Inter-annotator agreement is consistent across annotation tasks, particularly for identifying image-dependent questions, and indicates reliable application of the verification criteria. However, questions and filtering are fully automated using LLMs, with image metadata used to guide question generation, and human verification is not exhaustive. As a result, some visually subtle ambiguities or edge cases may persist in the full dataset.

**Binary judgment formulation:** M$^2$CQA formulates evaluation as independent true/false judgments. This enables precise contrastive analysis but does not capture graded uncertainty or partial visual support, which frequently occur in real-world multimodal reasoning. Models that adopt conservative rejection strategies may therefore appear more robust under CFHR despite reduced sensitivity to nuanced visual evidence.

**Evaluation-only focus:** This work is primarily diagnostic and does not propose training-time or inference-time methods to mitigate counterfactual hallucination. Although we show that prompting strategy significantly affects CFHR, we do not evaluate whether optimizing models against CFHR or incorporating M$^2$CQA during training improves visual grounding. Exploring mitigation strategies informed by conditional hallucination metrics is left to future work.

**Model and task scope:** Our experiments are limited to a specific set of contemporary vision–language models and to discriminative true/false reasoning. The extent to which CFHR generalizes to open-ended generation tasks such as image captioning, long-form visual reasoning, or multimodal dialogue remains an open question.

**Potential risks:** This work is intended as a diagnostic benchmark and poses minimal direct ethical or safety risks. However, culturally grounded counterfactual statements could be misinterpreted if taken out of context, potentially reinforcing stereotypes if treated as factual claims. In addition, the proposed CFHR metric may be misused if interpreted without regard to its conditional nature, for example by favoring overly conservative models. We mitigate these risks by framing M$^2$CQA and CFHR as evaluative tools rather than normative descriptions, by explicitly discussing proper interpretation of the metric, and by clearly stating the cultural and linguistic scope of the dataset.

## Ethics and Broader Impact Statement

The proposed dataset contains no personally identifiable information (PII). All images are obtained from publicly available sources, which were originally distributed under licenses that permit uses. The dataset is designed for multimodal hallucination analysis and does not intentionally include content targeting or disparaging any individual, group, organization, or community. Nonetheless, because publicly sourced imagery may incidentally contain sensitive or culturally contentious material, basic content screening has been done prior our data selection process. Annotations were annotated by trained annotators recruited through a third-party company and compensated at the standard hourly rate for their location. Annotators signed non-disclosure agreements prior to participation. We provide clear task instructions, avoid collecting annotator personal data beyond what is required for payment and administration by the vendor, and restrict dataset release to permitted uses consistent with the underlying licenses.

This dataset is intended to support research on detecting, measuring, and mitigating multimodal hallucinations, enabling more reliable vision language systems for downstream applications such as assistive QA. We recommend using it for evaluation and model development with explicit uncertainty reporting (e.g., abstention/calibration) and human-in-the-loop review in high-stakes settings.

# A Appendix

## A.1 Taxonomy

To summarize the semantic coverage of M$^2$CQA, we organize examples into a lightweight taxonomy (Table 3). The taxonomy provides an easy-to-read overview of the dataset and supports category-wise analysis of model behavior. It includes 12 top-level categories spanning built environments and landmarks, food and culinary culture, clothing and material culture, religion and national identity, everyday activities, transport and mobility, and environment/ecology, reflecting common visual themes in MENA imagery.

M$^2$CQA includes images collected from 17 countries across the MENA region. Figure 6 shows the proportion of visually grounded samples per country. In this work, we report results aggregated across countries, but the dataset enables future country-level analyses of model behavior.

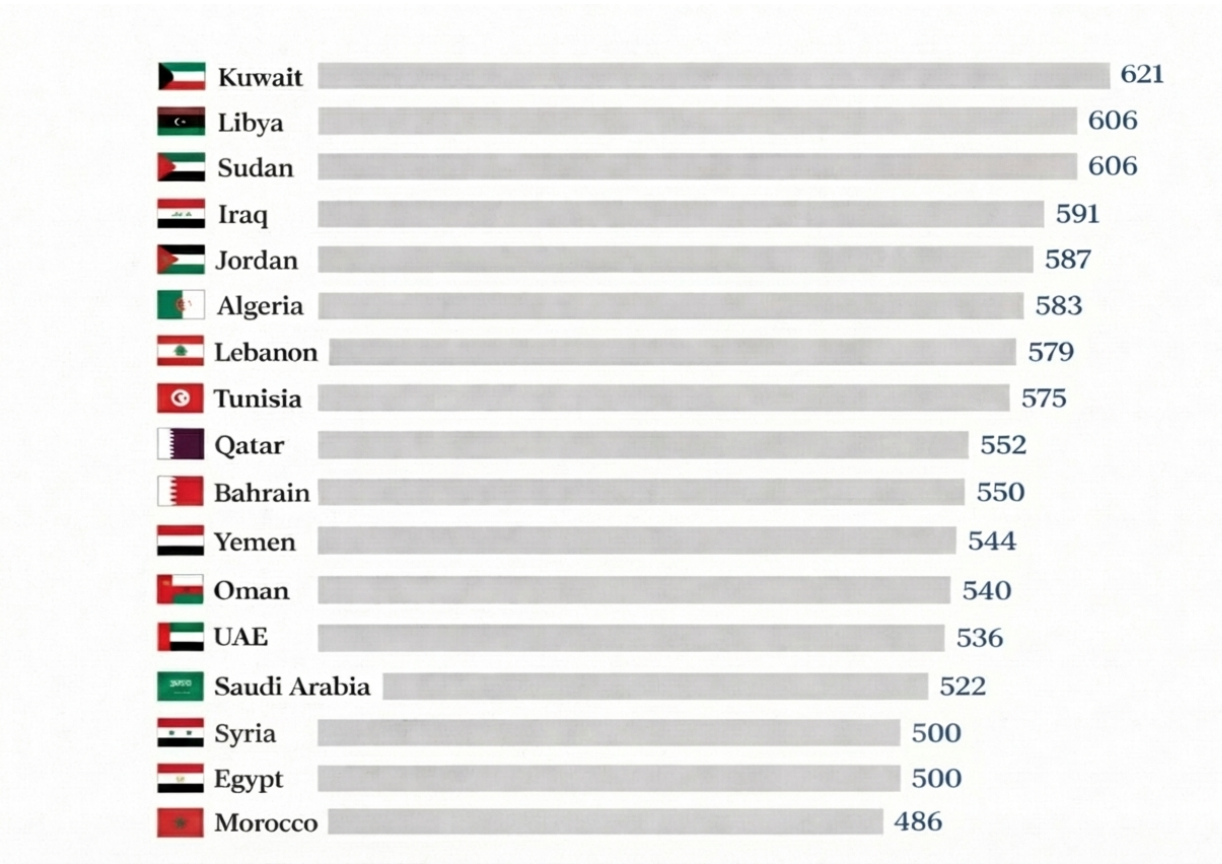

Figure 6: **Country coverage in M$^2$CQA.** Distribution of visually grounded samples across 17 MENA countries (total: 9,990).

## A.2 Counterfactual Curation and Validation

To curate counterfactual statements for M$^2$CQA, we adopt a two-stage annotation pipeline based on MCQs, which are used solely to enumerate culturally plausible interpretations of an image and are not used directly for model evaluation.

### A.2.1 Stage 1: Multiple-Choice Question Generation:

For each image, we generate a single MCQ consisting of exactly three answer options. One option corresponds to an interpretation that is visually supported by observable evidence in the image, while the remaining options are intended to be culturally plausible alternatives that are not supported by the visual content. The MCQ generation procedure is designed to emphasize reliance on image evidence and to reduce the influence of trivial or language-only cues. Answer options are written in a parallel and natural form, with distractors chosen to be plausible in isolation, such that identifying the correct option without access to the image is discouraged rather than guaranteed.

**MCQ curation Prompt:** We use the following system prompt in the first stage of dataset curation to generate MCQs. These MCQs serve as an intermediate annotation scaffold and are not used directly for evaluation.

```
You are an AI assistant specializing in
    Visual Question Answering (VQA).
Your task is to analyze the given image
    and generate a high-quality
multiple-choice question (MCQ) for
    benchmarking and training large
language models (LLMs).

Follow these guidelines carefully:

1. Question Type:
   - Generate exactly one multiple-
     choice question per image.
   - The question must have three
     plausible answer options.
   - Exactly one option must be correct
     and visually supported by the
     image.
   - The remaining options must be
     plausible but not supported by
     the image.

2. Semantic Focus:
   - Use the following semantic labels
     to guide question construction.
     Match the image content to the most
        relevant labels:
       * Location and Place
         Identification
       * Scene Interpretation and
         Context
       * Architectural Features and
         Functions
       * Cultural Significance and
         Heritage
       * Traditional Clothing and Attire
       * Tourism and Cultural Activities
       * Event and Activity Type
       * Objects, Animals, and Food
         Recognition
       * National Symbols and Identity
       * Visual Attributes
       * Recreational Activities and
         Facilities

3. Cognitive Focus:
   - Ensure that the question requires
     visual grounding in the image.
   - Assign a label indicating the
     cognitive focus:
```

| Category | Subcategories |
|---|---|
| **1. Built Environment & Architecture** | Modern landmarks; residential buildings; public spaces; historical/archaeological sites; government buildings; interior design; healthcare buildings; transport infrastructure; bridges; religious architecture. |
| **2. Food & Culinary Culture** | Prepared dishes; ingredients & herbs; beverages; tableware; food textures & garnish. |
| **3. Clothing, Textiles & Appearance** | Fabric materials; modern apparel; accessories; traditional/cultural clothing; patterns & motifs. |
| **4. Objects, Artifacts & Material Culture** | Materials; household objects; decorative crafts; technology devices; jewelry; furniture; machinery/tools; signage; religious objects. |
| **5. Nature, Animals & Ecology** | Animals; plants/vegetation; landscapes; water features; climate indicators. |
| **6. Activities, Sports & Human Action** | Competitive sports; motorsports; recreational activities; labor/work actions; festivals/performances; marketplace activities. |
| **7. Religion, Culture & National Identity** | Religious buildings; weddings; cultural celebrations; fashion events; political events; national symbols. |
| **8. Transport & Mobility** | Road transport; rail transport; air transport; water transport; animal-based transport. |
| **9. People & Occupations** | Everyday people; healthcare workers; manual laborers; performers; transport operators; athletes; religious/cultural participants. |
| **10. Environment, Climate & Geography** | Beaches/coasts; deserts/rocky terrains; forests; urban outdoors; disaster zones; rural countryside; lighting/weather conditions. |
| **11. Arts, Entertainment & Media** | Performing arts; film/media; amusement/festival structures; gaming/esports; fine arts/sculpture; museum/exhibition spaces; manuscripts/texts. |
| **12. Commerce, Markets & Economic Life** | Local markets; shopping centers; local goods; tourism commerce; restaurants/cafés; fairs/bazaars. |

Table 3: Hierarchical taxonomy of semantic categories for $M^2$CQA / ArabicMENA multimodal dataset.

```
      * knowledge-based
      * common sense-based

4. Language:
   - The question and answer options
       must be written in native-
       sounding
     English.

5. Question Quality:
   - Ensure the question is natural,
       conversational, and human-like.
   - Avoid trivial or language-only cues
        that would reveal the correct
     answer without access to the image.
   - Answer options should be written in
        a parallel and natural form.

6. Answer Quality:
   - The correct answer must be
       factually supported by the image.
   - Incorrect options must remain
       plausible in isolation.

7. Cultural Sensitivity:
   - Avoid stereotypes or cultural
       misrepresentations.
   - Ensure cultural references are
       accurate and specific to the
       image.

8. Context Utilization:
   - Use the provided image description,
        category, and subcategory to
     enrich question construction
          without making the answer
          obvious.

9. Reasoning:
   - Provide a short rationale (under
       100 words) explaining why the
     correct option is supported by the
         image and why the alternatives
     are not.

Output Format (JSON):
{
  "multiple-choice": {
    "question_en": "...",
    "options_en": ["...", "...", "..."],
    "correct_answer_en": "...",
    "rationale": "...",
    "cognitive_focus": "...",
    "semantic_focus": ["...", "..."]
  }
}
```

#### A.2.2 Stage 2: Contrastive Statement Construction:

Each MCQ is subsequently converted into a contrastive set of declarative statements. The correct MCQ option is rewritten as a standalone true statement ($Q^{+}$), while the two incorrect options are rewritten as false but culturally plausible counterfactual statements ($Q^{-}$). All statements are formulated as declarative sentences referring to the image and preserve the semantic content of their corresponding MCQ options, while removing explicit question structure. The resulting statements are constructed to be minimally contrastive, typically differing along fine-grained visual or cultural dimensions such as landmark identity, garment type, architectural feature, event category, or national association. Consistent with this contrastive design, explicit negation is avoided, and counterfactual statements are derived without introducing additional entities or relying on external knowledge beyond what is implied by the original MCQ options.

**Counterfactual Curation Prompt:** We use the following system prompt to convert each MCQ into a contrastive true/false format. This prompt performs only a deterministic conversion of MCQ answer options into declarative statements and does not introduce new content.

```
You are an AI assistant.

Your task is to convert a filtered
    multiple-choice question (MCQ)
into a set of declarative true and false
     statements.

Input:
- One multiple-choice question (MCQ)
- Three answer options
- The correct answer

Instructions:

1. Rewrite each answer option as a
    standalone declarative statement
   referring to the image.

2. From this conversion, produce:
   - One TRUE statement (Q+), derived
       from the correct answer option.
   - Two FALSE statements (Q-), derived
       from the incorrect answer options
       .

3. Preserve the semantic content of each
     answer option.
4. Remove explicit question structure.
5. Do not introduce new entities or
    additional information.

Output Format (JSON):
{
  "Q_plus": "...",
  "Q_minus": ["...", "..."]
}
```

### A.2.3 Verification of Minimal Contrastiveness:

We empirically assess the minimal contrastiveness of $Q^+$ and $Q^-$ statements along two dimensions. First, we examine base accuracy patterns and observe that $Q^-$ accuracy closely tracks $Q^+$ accuracy across models and languages (Table 2), suggesting that negative instances are not systematically more difficult. Second, we quantify semantic proximity between paired statements. We compute embeddings for (i) full true/false statements and (ii) answer choicesusing `all-MiniLM-L6-v2`, and measure cosine similarity. Intra-question similarity ($Q^+$ vs. paired $Q^-$) is substantially higher than inter-question baselines (0.71 vs. 0.19 for full statements; 0.39 vs. 0.16 for answer choices), indicating that counterfactual statements remain semantically close to their corresponding $Q^+$ instances. Together, these results validate the intended contrastive structure of the dataset.

### A.2.4 Number of Counterfactuals

CFHR conditions on correctly rejecting all counterfactual ($Q^-$) instances associated with a given image. As a result, the absolute magnitude of CFHR depends on the number of $Q^-$ statements: increasing the number of negatives lowers the probability of satisfying all constraints and therefore increases CFHR values. In preliminary experiments, we evaluated CFHR using a single $Q^-$ per image and observed consistent qualitative trends across models, languages, and prompting strategies. While absolute CFHR values differed due to the reduced number of constraints, relative comparisons remained stable. These observations suggest that CFHR is robust for comparative analysis, although its absolute scale depends on the number of counterfactual instances per image.

## A.3 Human Verification

Figure 12 shows the annotation interface used for human verification. Annotators are presented with a single image and a set of associated question–answer (Q/A) pairs, each evaluated independently along two dimensions: *(i)* **Q/A correctness**, indicating whether the answer is supported by observable visual evidence in the image (*Correct*, *Incorrect*, *Unsure*), and *(ii)* **Need-Image**, indicating whether answering the question requires access to the image (*Yes*, *No*, *Unsure*). These labels are defined independently: a Q/A pair may be visually grounded yet incorrect, or correct but answerable without access to the image.

Annotators are instructed to mark answers as *Incorrect* when visual evidence is absent, contradictory, irrelevant, or requires external knowledge not directly inferable from the image. The *Unsure* label is reserved for genuine visual ambiguity (e.g., blur, occlusion, or insufficient resolution), and is discouraged otherwise. The example in Figure 12 illustrates a grounded true statement ($Q^+$) alongside plausible but visually unsupported counterfactuals ($Q^-$), enabling fine-grained assessment of visual grounding.

| Statement | Need-Image | | Q-Correctness | |
|---|---|---|---|---|
| | AC1 | Raw | AC1 | Raw |
| $Q^+$ | 0.996 | 0.996 | 0.752 | 0.784 |
| $Q_1^-$ | 0.997 | 0.997 | 0.775 | 0.803 |
| $Q_2^-$ | 0.996 | 0.996 | 0.695 | 0.738 |
| Avg | 0.996 | 0.996 | 0.741 | 0.775 |

Table 4: Inter-annotator agreement for each annotation task and statement type.

We annotate[4] a stratified subset of approximately 15% of the dataset to ensure balanced coverage across countries and semantic categories. Each item is annotated by two annotators, with disagreements resolved by a third. The resulting subset spans all major domains defined by our taxonomy (Appendix A.1), covering diverse visual and cultural contexts. Category-level analysis further confirms broad coverage, with the largest groups including Objects & Material Culture (27.9%), Activities & Human Action (13.5%), and Clothing & Textiles (10.4%), alongside smaller but consistent representation across remaining categories.

We measure inter-annotator agreement using **Gwets AC1** (Gwet, 2008) and report raw percent agreement. Agreement is near-perfect for *Need-Image* (AC1 $\approx$ 0.996; raw $\approx$ 0.996), and substantial for *Q-Correctness* (AC1 $\approx$ 0.741; raw $\approx$ 0.775), consistent with the finer-grained judgments required for contrastive evaluation. Bootstrap 95% confidence intervals (item-level resampling) are strictly above zero across all conditions.

Qualitative inspection indicates that disagreements primarily arise from visually subtle counterfactual distinctions or fine-grained cultural ambiguity, rather than systematic annotation errors. Instances of image insufficiency are rare, consistent with the high agreement for *Need-Image*.

### A.4 Translation Quality

We assess translation quality using a combination of automatic and human evaluation to ensure that both MSA and dialectal Arabic variants are suitable for use in our multilingual and multimodal experiments.

[4] Annotators were recruited through a third-party provider, compensated at standard hourly rates, and are fluent in both Arabic and English. All annotators signed non-disclosure agreements (NDAs).

| Test Set | BLEU |
|---|---|
| **Levantine Arabic** | |
| MADAR test (LEV-0) | 20.25 |
| MADAR test (LEV-0-LB) | 12.10 |
| LDC test | 6.11 |
| MADAR test (LEV-1-JO) | 16.78 |
| **Average** | **13.81** |
| **Egyptian Arabic** | |
| MADAR test (NIL-2-EG) | 19.68 |
| ARZEn | 6.85 |
| MADAR test (NIL-1-EG) | 18.63 |
| MADAR test (NIL-0-EG) | 18.12 |
| **Average** | **15.82** |

Table 5: BLEU scores for direct English-to-dialect Arabic translation using GPT-4.1 on standard dialectal test sets.

#### A.4.1 Automatic Evaluation

As part of the data construction pipeline, MSA variants are produced using the in-house **Fanar Shaheen** machine translation system (Team et al., 2026), while dialectal Arabic variants are generated via direct English-to-dialect translation using GPT-4.1. To assess the quality of dialectal translations, we conduct automatic evaluations using BLEU on standard dialectal test sets.

Table 5 reports BLEU scores for direct English-to-dialect translation across Levantine and Egyptian Arabic. Performance varies across test sets, reflecting known differences in domain, orthographic conventions, and the inherent difficulty of dialectal Arabic translation. While BLEU scores are modest, they are consistent with prior observations in the literature and indicate that GPT-4.1 produces translations of sufficient quality for downstream use in our evaluation pipeline.

#### A.4.2 Human Evaluation

To complement automatic metrics, we conduct a targeted human evaluation to directly assess translation adequacy and fluency in the context of our task. We randomly sample 100 multiple-choice questions per language variant and ask native or fluent Arabic speakers to rate the quality of translated questions on a 6-point scale ranging from 0 (completely incorrect or unusable) to 5 (excellent, accurate, and natural). We focus our primary analysis on translated questions rather than answer options, as questions are longer and structurally

| Variant | Mean | Median | $\geq 4(\%)$ |
|---|---|---|---|
| MSA | 4.66 | 5.0 | 90% |
| ARZ | 4.41 | 4.5 | 94% |
| AJP | 4.36 | 4.0 | 87% |

Table 6: Human evaluation of question translation quality (mean score on a 0–5 scale). ARZ: Egyptian Arabic; AJP: Levantine Arabic.

more complex, making them a more conservative and informative measure of translation quality.

Table 6 summarizes the results. Translations across all varieties achieve high average scores, with mean ratings of 4.66 for MSA, 4.41 for Egyptian Arabic, and 4.36 for Levantine Arabic. More than 87% of translations in all varieties are rated as Good or Excellent (4), and fewer than 2% are rated below Fair (2). These results indicate that translation errors are rare and unlikely to meaningfully affect downstream evaluation.

### A.5 Prompting Strategies

We evaluated models using three fixed prompt templates (Table 8) that differ in the timing and structure of reasoning relative to the final decision. **Prompt A (True or False)** requires the model to directly predict whether a statement is true or false with respect to the image, without providing any justification. **Prompt B (Answer then Reason)** requires the model to first commit to a binary decision and then provide supporting evidence grounded in the image. **Prompt C (Reasoning First)** reverses this order by eliciting explicit reasoning steps prior to the final true/false decision. Each prompt is instantiated in four language variants: English (EN), Modern Standard Arabic (MSA), Levantine Arabic (AJP), and Egyptian Arabic (ARZ), using semantically equivalent instructions and strictly enforced output formats to ensure consistent parsing across models and languages. For all settings, only the final binary decision is used for scoring; any output that does not match the required format is treated as incorrect. This controlled prompting setup allows us to isolate the effect of reasoning order and linguistic variation on counterfactual hallucination behavior.

**Statistical Significance Analysis** To assess whether prompt-level differences in CFHR are robust, we conduct non-parametric bootstrap significance testing using 1,000 iterations with item-level sampling and replacement. Table 7 reports the mean differences ($\Delta$), 95% confidence intervals, and two-sided p-values for all prompt-pair comparisons across models and languages. Across models and languages, nearly all differences between prompting strategies are statistically significant (p $< 0.001$), with confidence intervals excluding zero. The only exception is Fanar in English for Prompt A vs. Prompt B ($\Delta = 0.002$), where the confidence interval includes zero and the result is not statistically significant ($p = 0.11$). These results confirm that the observed prompt-level effects are robust rather than sampling artifacts.

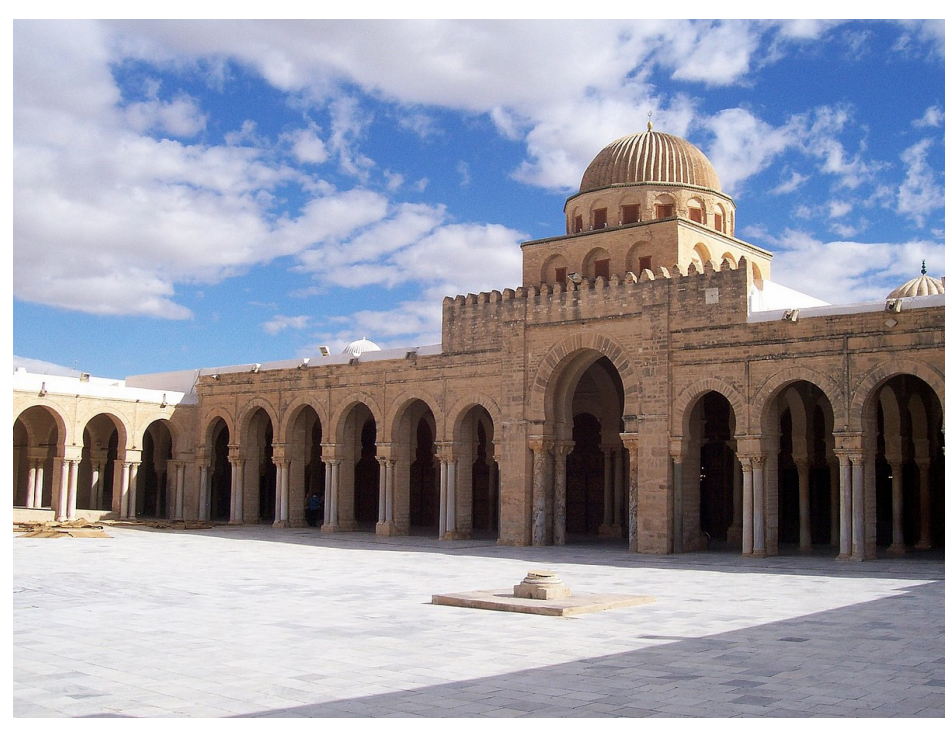

Figure 7: Example image used to illustrate the prompt templates in Table 8.

### A.6 Qualitative Analysis of Reasoning-First Prompting

Table 9 provides qualitative examples that illustrate the mechanisms underlying the quantitative trends reported in RQ4. In all three cases, the model produces the ground-truth judgment under *Prompt B* (*Answer then Reason*), but reverses its decision under *Prompt C* (*Reasoning First*), despite correctly perceiving the relevant visual evidence in both settings. These examples highlight how eliciting reasoning prior to commitment can introduce additional constraints that are not required by the task and are not supported by the image.

In the **doll example**, the model under Prompt C correctly identifies the presence of a metallic pitcher and an orange-colored component, but subsequently reframes the task by imposing a stricter semantic interpretation of object individuation. By reasoning that the orange element is attached to the metallic pitcher and therefore does not constitute a separate object, the model rejects the statement despite its visual correctness. This failure does not arise from misperception, but from an unnecessary decomposition of the scene introduced during reasoning.

| Prompt A vs B | | | | |
|---|---|---|---|---|
| **Lang** | **Model** | $\Delta$ | **CI** | $p$ |
| en | Qwen3-VL-32B-Instruct | 0.018 | [0.016, 0.022] | $< 0.001$ |
| en | gemma-3-27b-it | 0.010 | [0.006, 0.014] | $< 0.001$ |
| en | FanarOryx-v1 | 0.002 | [-0.004, 0.001] | $= 0.11$ |
| en | AIN | 0.031 | [0.027, 0.035] | $< 0.001$ |
| msa | Qwen3-VL-32B-Instruct | 0.023 | [0.019, 0.026] | $< 0.001$ |
| msa | gemma-3-27b-it | 0.028 | [0.023, 0.032] | $< 0.001$ |
| msa | FanarOryx-v1 | 0.497 | [0.487, 0.509] | $< 0.001$ |
| msa | AIN | 0.199 | [0.189, 0.208] | $< 0.001$ |

| Prompt B vs C | | | | |
|---|---|---|---|---|
| **Lang** | **Model** | $\Delta$ | **CI** | $p$ |
| en | Qwen3-VL-32B-Instruct | 0.042 | [-0.047, -0.037] | $< 0.001$ |
| en | gemma-3-27b-it | 0.017 | [0.011, 0.022] | $< 0.001$ |
| en | FanarOryx-v1 | 0.009 | [-0.015, -0.003] | $< 0.001$ |
| en | AIN | 0.146 | [-0.155, -0.137] | $< 0.001$ |
| msa | Qwen3-VL-32B-Instruct | 0.031 | [-0.037, -0.026] | $< 0.001$ |
| msa | gemma-3-27b-it | 0.048 | [0.041, 0.055] | $< 0.001$ |
| msa | FanarOryx-v1 | 0.228 | [-0.239, -0.217] | $< 0.001$ |
| msa | AIN | 0.310 | [-0.323, -0.297] | $< 0.001$ |

| Prompt A vs C | | | | |
|---|---|---|---|---|
| **Lang** | **Model** | $\Delta$ | **CI** | $p$ |
| en | Qwen3-VL-32B-Instruct | 0.023 | [-0.028, -0.019] | $< 0.001$ |
| en | gemma-3-27b-it | 0.027 | [0.021, 0.033] | $< 0.001$ |
| en | FanarOryx-v1 | 0.011 | [-0.017, -0.005] | $= 0.001$ |
| en | AIN | 0.115 | [-0.124, -0.105] | $< 0.001$ |
| msa | Qwen3-VL-32B-Instruct | 0.009 | [-0.014, -0.003] | $< 0.001$ |
| msa | gemma-3-27b-it | 0.076 | [0.069, 0.083] | $< 0.001$ |
| msa | FanarOryx-v1 | 0.269 | [0.255, 0.283] | $< 0.001$ |
| msa | AIN | 0.111 | [-0.125, -0.097] | $< 0.001$ |

Table 7: Statistical significance of CFHR differences across prompting strategies using bootstrap resampling (1,000 iterations). We report mean differences ($\Delta$), 95% confidence intervals (CI), and two-sided p-values. Nearly all comparisons are statistically significant, with the exception of Fanar in English for Prompt A vs. Prompt B.

In the **biryani example**, the model under Prompt C accurately recognizes star anise in the image, yet overrides this visual evidence by appealing to background knowledge about regional cuisine. The model reframes the statement as a claim about culinary typicality rather than visual presence, leading it to reject an otherwise correct answer. This illustrates how reasoning-first prompting can encourage the model to privilege general world knowledge over image-grounded evidence.

In the **lantern example**, the model under Prompt C acknowledges the presence of colorful hanging lanterns, but shifts the decision criterion from visual salience to architectural dominance. By reinterpreting most prominent in terms of structural surface area rather than visual emphasis, the model again introduces an additional constraint that is not specified by the question, resulting in a counterfactual rejection.

Across all three examples, a common pattern emerges: reasoning-first prompting encourages the model to explore alternative interpretations and to refine the decision criterion beyond what the task requires. While such reasoning is internally coherent, it is not anchored to the evaluation objective, and it often leads to the rejection of visually supported statements. In contrast, committing to an answer prior to justification appears to anchor the models decision to image-grounded evidence, with subsequent reasoning serving primarily to rationalize rather than reinterpret the judgment.

These qualitative findings support the quantitative results in RQ4, demonstrating that the timing of evidence elicitation plays a critical role in counterfactual hallucination. Reasoning is beneficial only when it follows an initial commitment; when elicited beforehand, it can systematically amplify semantic over-interpretation and reliance on cultural or conceptual priors at the expense of visual grounding.

| | EN | MSA | AJP | ARZ |
|---|---|---|---|---|
| **A** | Your task is to decide whether the following statement is True or False. Please respond exactly in the format of ‘The final answer is: <True/False>’.<br>Statement: The most prominent feature of the mosque in the image is a tall minaret<br>Answer: False | مهمتك هي تحديد ما إذا كانت العبارة التالية صحيحة أم خاطئة. يُرجى الإجابة بصيغة `الإجابة النهائية هي: <صحيح/خطأ>'.<br>العبارة: الميزة الأهم في المسجد في الصورة هي مئذنة طويلة<br>الجواب: خطأ | مهمتك هي تحديد ما إذا كانت العبارة التالية صحيحة أم خاطئة. يُرجى الإجابة بصيغة `الإجابة النهائية هي: <صحيح/خطأ>'.<br>العبارة: هالمسجد اللي بالصورة فيه مئذنة طويلة<br>الجواب: غلط | مهمتك هي تحديد ما إذا كانت العبارة التالية صحيحة أم خاطئة. يُرجى الإجابة بصيغة `الإجابة النهائية هي: <صحيح/خطأ>'.<br>العبارة: المئذنة الطويلة هي أكتر حاجة باينة في الجامع اللي في الصورة<br>الجواب: غلط |
| **B** | Your task is to decide whether the following statement is True or False, and then provide a justification. Please respond exactly in the format of ‘The final answer is: <True/False>. Evidence: <your evidence>.’.<br>Statement: The most prominent feature of the mosque in the image is a tall minaret<br>Answer: False | مهمتك هي تحديد ما إذا كانت العبارة التالية صحيحة أم خاطئة، ثم تقديم مبرر لقرارك. يُرجى الإجابة بصيغة `الإجابة النهائية هي: <صحيح/خطأ>. الدليل: <دليلك>'.<br>العبارة: الميزة الأهم في المسجد في الصورة هي مئذنة طويلة<br>الجواب: خطأ | مهمتك هي تحديد ما إذا كانت العبارة التالية صحيحة أم خاطئة، ثم تقديم مبرر لقرارك. يُرجى الإجابة بصيغة `الإجابة النهائية هي: <صحيح/خطأ>. الدليل: <دليلك>'.<br>العبارة: هالمسجد اللي بالصورة فيه مئذنة طويلة<br>الجواب: غلط | مهمتك هي تحديد ما إذا كانت العبارة التالية صحيحة أم خاطئة، ثم تقديم مبرر لقرارك. يُرجى الإجابة بصيغة `الإجابة النهائية هي: <صحيح/خطأ>. الدليل: <دليلك>'.<br>العبارة: المئذنة الطويلة هي أكتر حاجة باينة في الجامع اللي في الصورة<br>الجواب: غلط |
| **C** | Your task is to decide whether the following statement is True or False. Please think about the statement and provide your thinking steps. Please respond exactly in the format of ‘Thinking Steps: <thinking_steps>. The final answer is <True/False>’.<br>Statement: The most prominent feature of the mosque in the image is a tall minaret<br>Answer: False | مهمتك هي تحديد ما إذا كانت العبارة التالية صحيحة أم خاطئة. يرجى التفكير في العبارة وتقديم خطوات تفكيرك. يُرجى الإجابة بصيغة `خطوات التفكير: <خطوات_التفكير>. الإجابة النهائية هي <صحيح/خطأ>'.<br>العبارة: الميزة الأهم في المسجد في الصورة هي مئذنة طويلة<br>الجواب: خطأ | مهمتك هي تحديد ما إذا كانت العبارة التالية صحيحة أم خاطئة. يرجى التفكير في العبارة وتقديم خطوات تفكيرك. يُرجى الإجابة بصيغة `خطوات التفكير: <خطوات_التفكير>. الإجابة النهائية هي <صحيح/خطأ>'.<br>العبارة: هالمسجد اللي بالصورة فيه مئذنة طويلة<br>الجواب: غلط | مهمتك هي تحديد ما إذا كانت العبارة التالية صحيحة أم خاطئة. يرجى التفكير في العبارة وتقديم خطوات تفكيرك. يُرجى الإجابة بصيغة `خطوات التفكير: <خطوات_التفكير>. الإجابة النهائية هي <صحيح/خطأ>'.<br>العبارة: المئذنة الطويلة هي أكتر حاجة باينة في الجامع اللي في الصورة<br>الجواب: غلط |

Table 8: Prompt templates with ground-truth answers in English and Arabic variants, shown using the example image in Figure 7.

### A.7 Effect of Scaling on Hallucination

In the main paper (Figure 5), we showed that counterfactual hallucination generally decreases as model size increases under the direct true/false setting (Prompt A), with Qwen3-VL benefiting more from scaling than Gemma-3-VL, particularly in Arabic. Figures 8 in the appendix show the corresponding results for Prompt B (Answer then Reason) and Prompt C (Reasoning First).

Across both prompts, we observe the same overall pattern: larger models tend to hallucinate less, scaling effects are stronger in MSA and dialectal Arabic than in English, and Qwen3-VL shows more consistent gains from increased model size than Gemma-3-VL. While Prompt C leads to higher hallucination rates overall, especially for smaller models, scaling still results in clear reductions in CFHR. Together, these results show that the scaling trends reported in the main paper hold across different prompting and reasoning setups.

In Figures 9, we report scaling results separately for Egyptian (AJP) and Levantine (ARZ) Arabic under Prompts A–C. Across all prompts, counterfactual hallucination generally decreases with model size in both dialects, with stronger and more consistent gains for Qwen3-VL than for Gemma-3-VL. For Qwen, CFHR drops steadily as scale increases and the gap between AJP and ARZ narrows at larger sizes, indicating improved robustness to dialectal variation. In contrast, Gemma-3-VL shows persistently higher CFHR with limited scal-

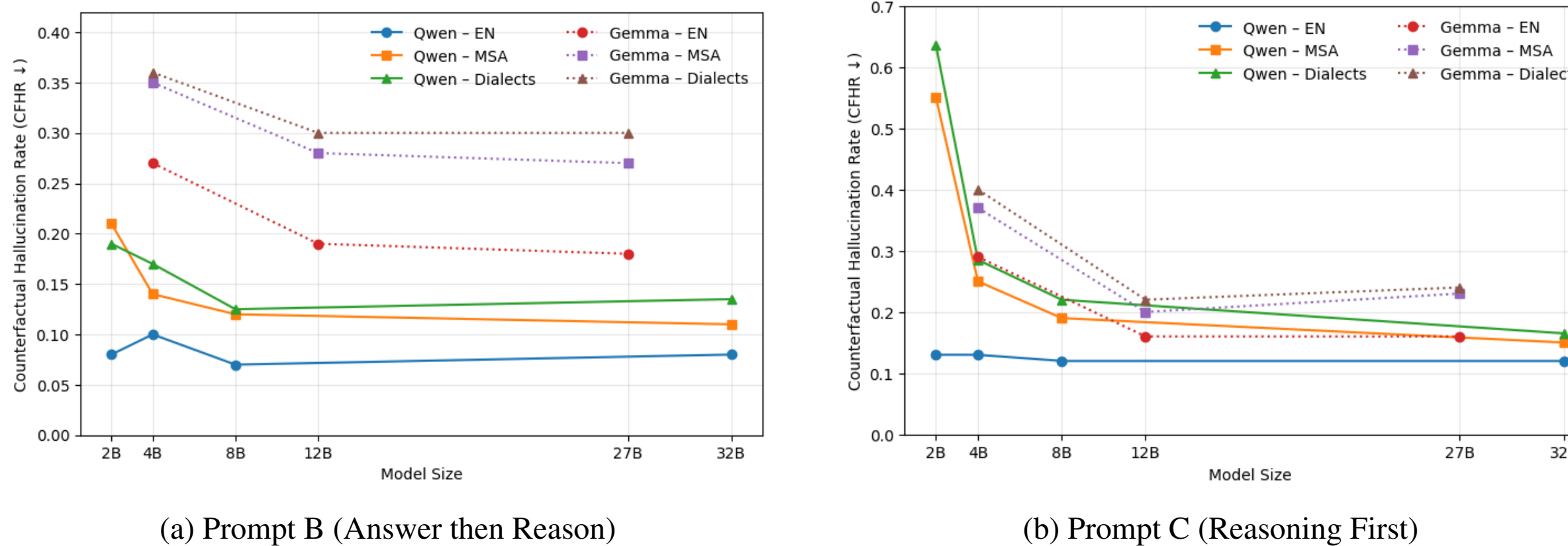

(a) Prompt B (Answer then Reason)

(b) Prompt C (Reasoning First)

Figure 8: **Scaling and counterfactual hallucination across prompting strategies.** Counterfactual Hallucination Rate (CFHR; lower is better) as a function of model size for Qwen3-VL (solid) and Gemma-3-VL (dotted), evaluated in English (EN), Modern Standard Arabic (MSA), and dialectal Arabic (Dialects; average of Egyptian and Levantine). Prompt B requires models to answer before providing justification, while Prompt C elicits reasoning prior to the final answer.

ing benefits and near-identical behavior across the two dialects. Prompt B consistently reduces hallucination relative to Prompt A, while Prompt C increases CFHR at smaller scales, though scaling still yields clear improvements. Overall, these results confirm that the scaling trends reported in the main paper for averaged dialects also hold at the level of individual Arabic dialects.

### A.8 Sample Images

Figures 10 and 11 show representative samples from the $M^2$CQA dataset, illustrating the breadth of visual content and the contrastive question design used throughout the benchmark. Each image is paired with one visually grounded true statement ($Q^+$) and two culturally plausible but visually unsupported counterfactual statements ($Q^-$), presented in both English and MSA.

The examples span a wide range of everyday and culturally salient scenes, including objects, people, architecture, public spaces, religious artifacts, landmarks, marketplaces, and festive settings commonly encountered in the MENA region. Counterfactual questions are designed to be semantically and culturally reasonable while remaining verifiably incorrect given the visual evidence.

These sample images illustrate the central challenge targeted by $M^2$CQA: distinguishing visually grounded interpretations from culturally familiar but unsupported alternatives. They highlight why accurate visual grounding is necessary to avoid hallucinations driven by strong cultural or contextual priors, motivating the evaluation framework introduced in the main body.

## B Data Release

The dataset will be released under the CC BY-NC-SA 4.0 – Creative Commons Attribution 4.0 International License: [5].

[5] https://creativecommons.org/licenses/by-nc-sa/4.0/

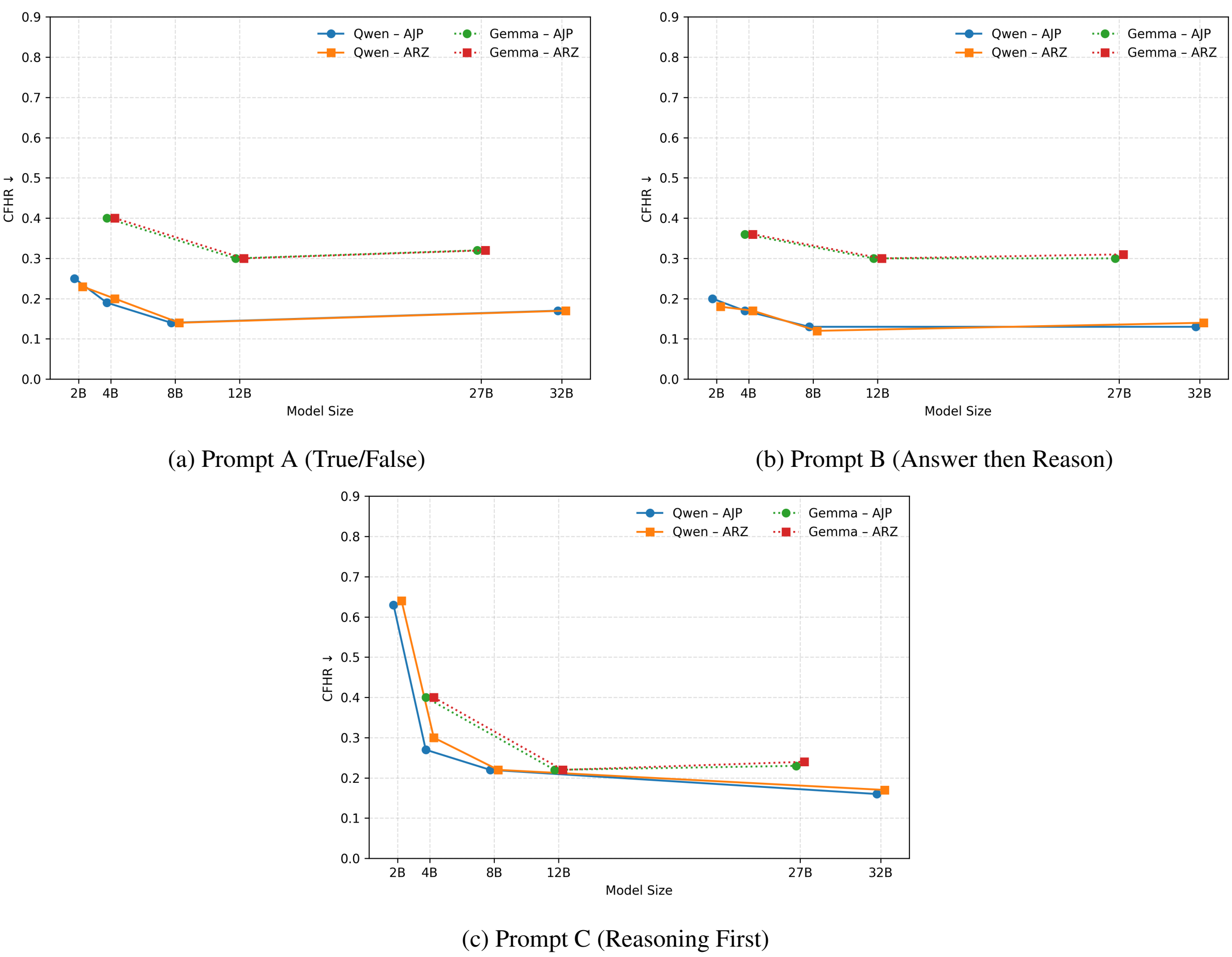

(a) Prompt A (True/False)

(b) Prompt B (Answer then Reason)

(c) Prompt C (Reasoning First)

Figure 9: **Dialectal scaling of counterfactual hallucination.** CFHR (lower is better) as a function of model size for Qwen3-VL (solid) and Gemma-3 (dotted), evaluated separately in Levantine Arabic (AJP) and Egyptian Arabic (ARZ).

| Image, Question & Ground Truth | Model Response (Prompt B) | Model Response (Prompt C) |
|---|---|---|
| 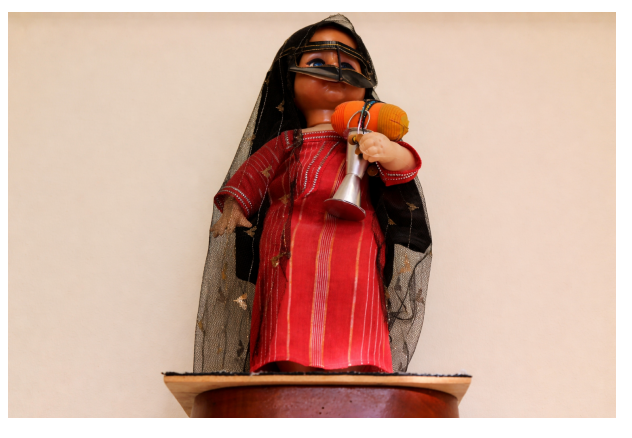<br>**Question:** True or False: The doll appears to be holding a metallic pitcher and an orange item in its hands.<br>**Ground Truth:** True | **The final answer is: True.**<br>**Evidence:** The doll is visibly holding a metallic pitcher, identifiable by its shiny, silver-colored surface and spout, in one hand, and an orange-colored item, which appears to be a knitted or crocheted object, in the other hand. Both items are clearly visible and held by the dolls hands. | **Reasoning:** The image shows a doll dressed in traditional attire, including a red garment and a black veil the doll is holding an object in its right hand that has a metallic, shiny appearance and a shape consistent with a small pitcher or jug additionally, the object has an orange-colored top or handle, which could be interpreted as an "orange item" the object appears to be a single item with both metallic and orange components, possibly a decorative or symbolic object rather than two separate items however, the statement says "a metallic pitcher and an orange item," which implies two distinct objects given that the orange part appears to be attached to the metallic pitcher (possibly a handle or lid), it is more accurate to describe it as one object with two features rather than two separate items therefore, the statement is misleading or inaccurate in implying two separate items.<br>**The final answer is: False.** |
| 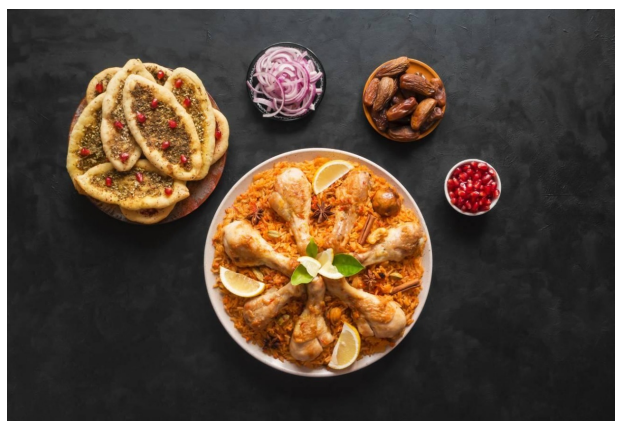<br>**Question:** Star anise is commonly used in Middle Eastern cuisine and is visible in the chicken biryani shown in the image. True or False?<br>**Ground Truth:** True | **The final answer is: True.**<br>**Evidence:** Star anise is a common spice in Middle Eastern and South Asian cuisines, often used in rice dishes like biryani for its distinctive licorice-like flavor. In the image, star anise pods are clearly visible among the rice and chicken in the central dish, confirming its presence in the chicken biryani. | **Reasoning:** The image shows a dish of chicken biryani, which is a rice dish commonly associated with South Asian cuisine, particularly Indian and Pakistani cooking. Star anise is a spice that is indeed used in some Middle Eastern dishes, but it is more prominently featured in South Asian and East Asian cuisines. In the image, star anise is visible as whole pods placed on top of the biryani, which is consistent with its use in biryani recipes. However, the statement claims that star anise is "commonly used in Middle Eastern cuisine," which is not entirely accurate. While star anise may appear in some Middle Eastern dishes, it is not a staple or commonly used spice in the region compared to its prevalence in South Asian cooking. Therefore, the statement is misleading due to the incorrect generalization about Middle Eastern cuisine.<br>**The final answer is: False.** |
| 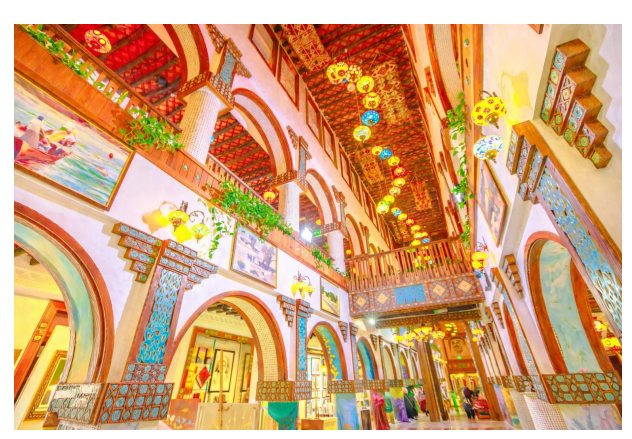<br>**Question:** The most prominent decorative element in the ceiling area of this heritage building is colorful hanging lanterns. True or False?<br>**Ground Truth:** True | **The final answer is: True.**<br>**Evidence:** The ceiling area of the building is adorned with numerous colorful hanging lanterns in various shapes and patterns, including red, yellow, blue, and floral designs, which are prominently displayed along the length of the ceiling and are the most visually striking decorative feature compared to the patterned wooden beams and other architectural elements. | **Reasoning:** The image shows an interior space with a high ceiling featuring a richly decorated wooden structure with red and gold patterns. While there are colorful hanging lanterns suspended from the ceiling, they are not the most prominent decorative element. The ceiling itself is adorned with intricate geometric and lattice patterns in red, gold, and dark wood, which cover a large surface area and dominate the visual field. The lanterns, though vibrant and numerous, are smaller in scale and distributed along the length of the ceiling, whereas the ceilings architectural design and patterns are more extensive and visually dominant. Therefore, the most prominent decorative element is the ceilings ornate wooden design, not the lanterns.<br>**The final answer is: False.** |

Table 9: Qualitative comparison of model outputs for Prompt B (*Answer then Reason*) and Prompt C (*Reasoning First*).

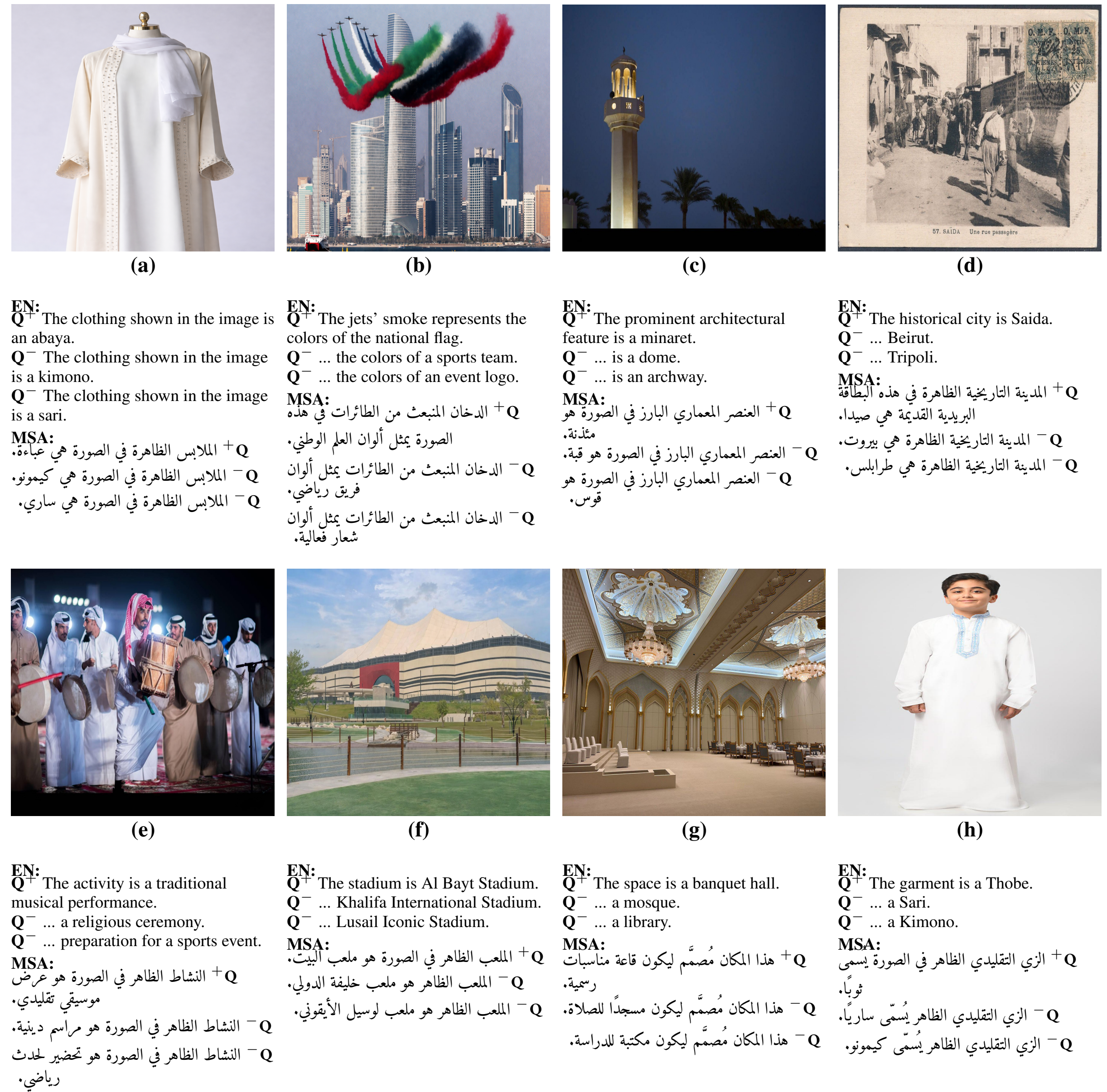

Figure 10: Sample images from the dataset with English and Modern Standard Arabic (MSA) captions.

| | | **A. True or False** | | | | **B. Answer then Reason** | | | | **C. Reasoning First** | |
|---|---|---|---|---|---|---|---|---|---|---|---|
| **Lang** | **Model** | $Q^+$ | $Q^-$- | F1 | **CFHR** ↓ | $Q^+$ | $Q^-$- | F1 | **CFHR** ↓ | **CFHR** ↓ | ΔCFHR |
| AJP | Qwen3-VL-32B | 0.89 | 0.89 | 0.89 | **0.17** | 0.86 | 0.92 | 0.89 | **0.13** | **0.16** | +0.03 |
| | Gemma-3-27B | 0.94 | 0.79 | 0.86 | **0.32** | 0.93 | 0.80 | 0.86 | **0.30** | **0.23** | −0.07 |
| | FanarOryx-v1 | 0.73 | 0.92 | 0.82 | **0.15** | 0.62 | 0.95 | 0.75 | **0.11** | **0.37** | +0.26 |
| | AIN | 0.96 | 0.41 | 0.58 | **0.76** | 0.89 | 0.75 | 0.81 | **0.39** | **0.63** | +0.24 |
| ARZ | Qwen3-VL-32B | 0.87 | 0.90 | 0.88 | **0.17** | 0.83 | 0.92 | 0.87 | **0.14** | **0.17** | +0.03 |
| | Gemma-3-27B | 0.93 | 0.79 | 0.86 | **0.32** | 0.92 | 0.81 | 0.86 | **0.31** | **0.24** | −0.07 |
| | FanarOryx-v1 | 0.68 | 0.92 | 0.79 | **0.15** | 0.56 | 0.95 | 0.70 | **0.11** | **0.40** | +0.29 |
| | AIN | 0.97 | 0.36 | 0.53 | **0.81** | 0.92 | 0.69 | 0.79 | **0.47** | **0.62** | +0.15 |

Table 10: Dialectal Arabic results reported separately for Levantine Arabic (AJP) and Egyptian Arabic (ARZ). Prompt A corresponds to *True or False*, Prompt B to *Answer then Reason*, and Prompt C to *Reasoning First*. CFHR is the CounterFactual Hallucination Rate (conditional hallucination). ΔCFHR is defined as CFHR(C) − CFHR(B).

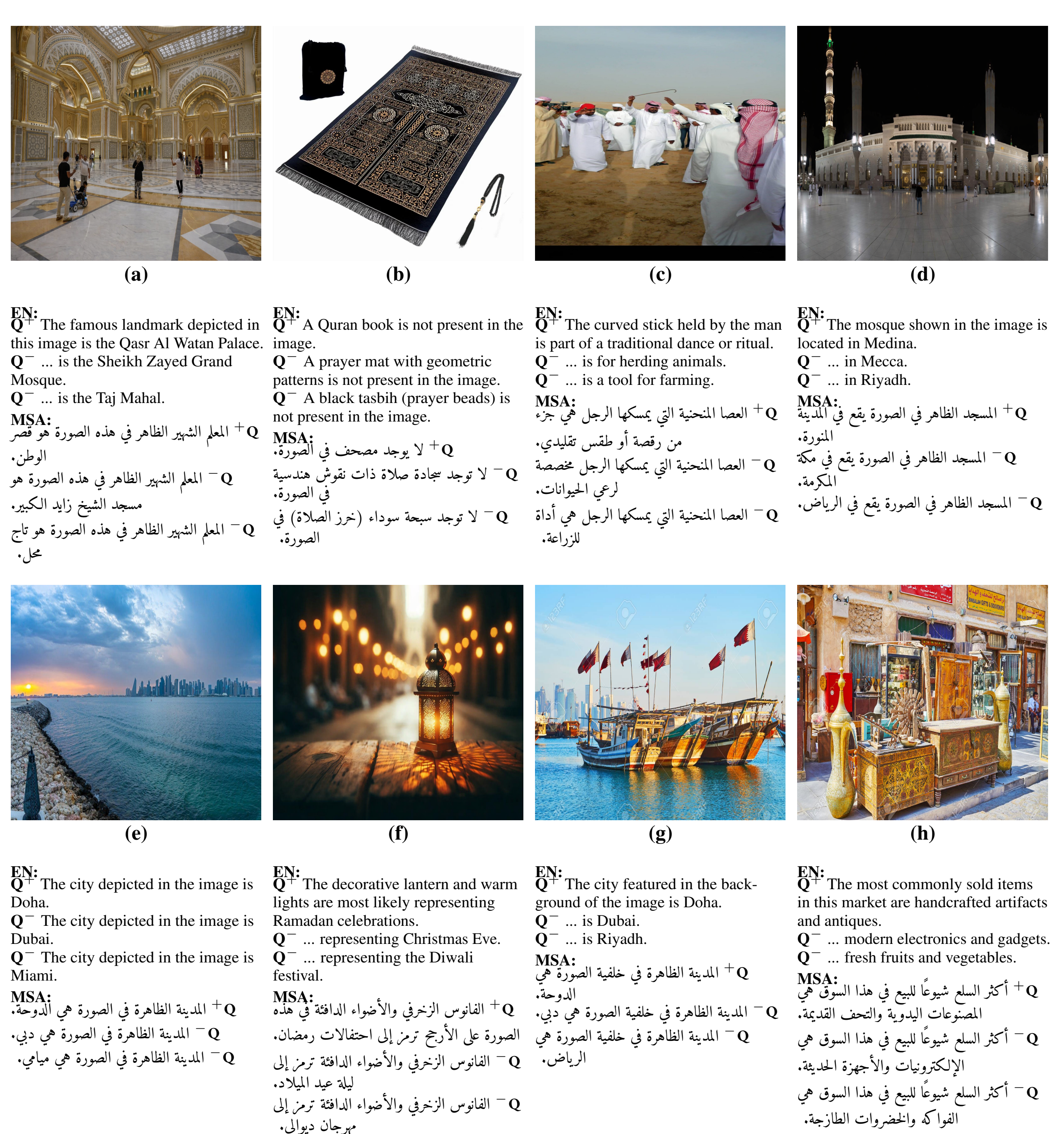

Figure 11: Sample images from the dataset with English and Modern Standard Arabic (MSA) captions.

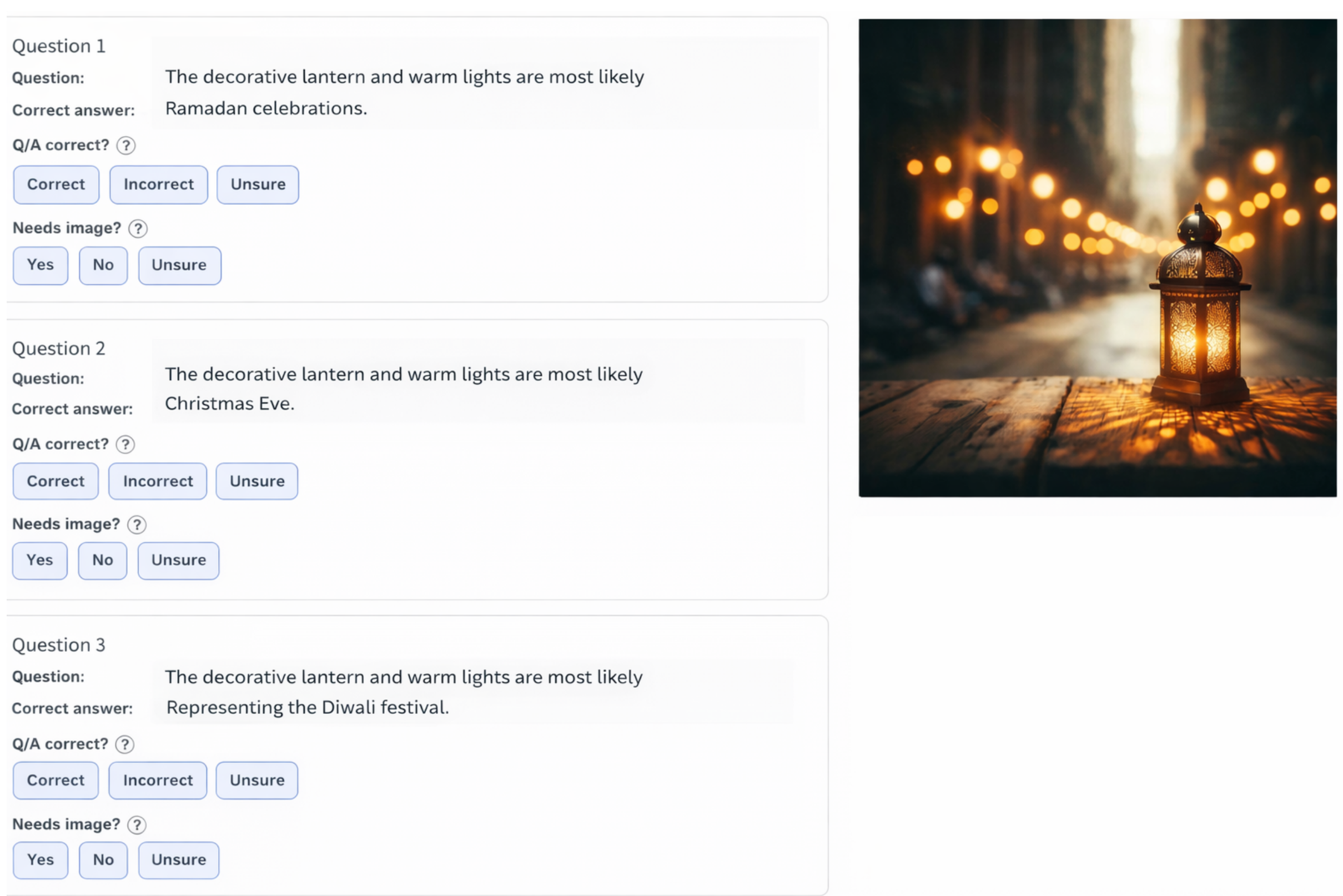

Figure 12: Human annotation interface used for image-based Q/A verification. Annotators evaluate each question–answer pair with respect to visual correctness and whether answering requires the image.